\documentclass{article}
\PassOptionsToPackage{table}{xcolor}
\usepackage{iclr2027_conference,times}

\usepackage[T1]{fontenc}
\usepackage[utf8]{inputenc}
\usepackage{hyperref}
\usepackage{url}
\usepackage{graphicx}
\usepackage{booktabs}
\usepackage{amsmath,amssymb}
\usepackage{cleveref}
\usepackage{placeins}
\usepackage{float}
\usepackage{capt-of}
\usepackage{wrapfig}

\iclrfinalcopy

\title{IECLA: Interpretable Emotion Contribution Learning for Affective Image Editing}

\author{\parbox{\dimexpr\textwidth-2\tabcolsep\relax}{\centering\vspace{10pt}
\textbf{Yijie Guo$^{1}$, Dexiang Hong$^{1}$, Weidong Chen$^{1}$\thanks{Corresponding author: \texttt{chenweidong@ustc.edu.cn}.}, Xinyan Liu$^{2}$, Zhendong Mao$^{1}$}\\[4pt]
{\small\normalfont $^{1}$University of Science and Technology of China\\
$^{2}$Harbin Institute of Technology}\\[3pt]
{\footnotesize\normalfont\texttt{\{guoyijie,hongdexiang\}@mail.ustc.edu.cn}\\
\texttt{\{chenweidong,zdmao\}@ustc.edu.cn}, \texttt{xinyliu@hit.edu.cn}}}}

\begin{document}

\maketitle
% Clear the conference-status header for the arXiv preprint.
\lhead{}

\vspace{-20pt}
\begin{center}
\includegraphics[width=\textwidth,pagebox=cropbox]{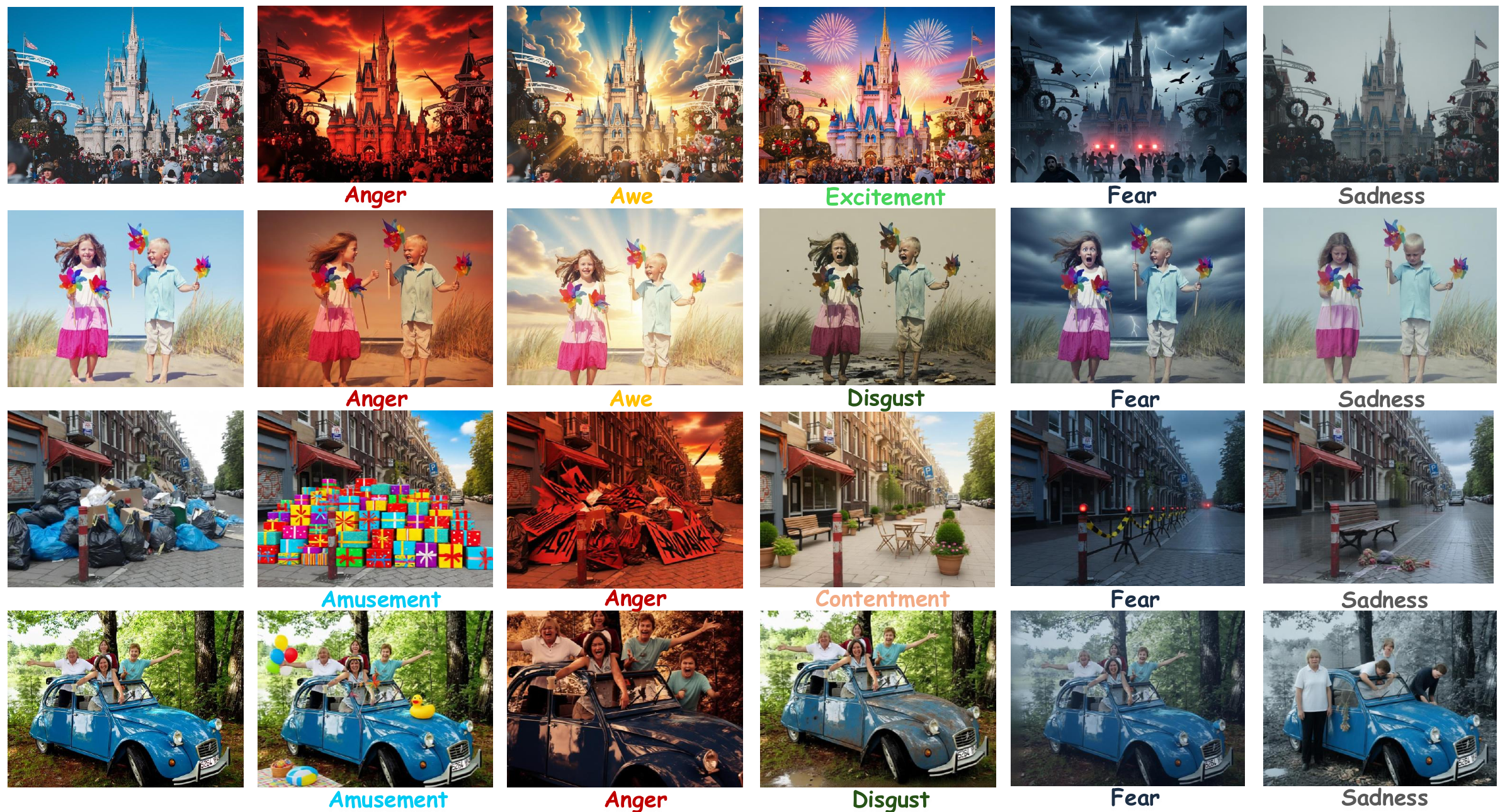}
\vspace{-16pt}
\captionof{figure}{Affective image editing with IECLA across diverse scenes and target emotions. Given only a target-emotion word, IECLA produces edits that effectively evoke the requested emotion while preserving the source image's content and structure.}
\label{fig:teaser}
\end{center}

\begin{abstract}
Affective image editing (AIE) aims to evoke a target emotion while preserving the content and structure of a source image. Existing methods improve emotional expression or visual fidelity, but the target emotion alone does not reveal which source regions should be changed and which should remain intact, especially in complex scenes. We introduce Interpretable Emotion Contribution Learning for Affective Image Editing (IECLA), a framework that separates source-emotion attribution from target-aware editing. IECLA constructs signed object-level supervision through counterfactual interventions and trains a context-conditioned contribution predictor to localize evidence that supports or opposes the source emotion. At inference, an image-level evaluator estimates the source emotion directly from the input, while the predictor identifies editable and preservation evidence without repeated interventions. A target-aware editing agent then translates this evidence and the desired emotion into localized, content-aware operations for existing editors. Extensive evaluations show that IECLA predicts counterfactual contributions more faithfully on both in-domain and cross-dataset tests, while external VLM and independent human references support the quality of its localized evidence. Across six editing backbones, IECLA consistently strengthens target-emotion expression while preserving image content and structure, and user studies further show improved perceived target expression and naturalness.\footnote{Code will be released in the final version of the paper.}
\end{abstract}

\section{Introduction}
\label{sec:introduction}

Affective image editing (AIE) seeks to evoke a target emotion while preserving the source image's identity and structure. This objective differs from an attribute-level instruction: an emotion word specifies a perceptual outcome, but does not indicate which visual evidence should change. The relevant cues may lie in a person's expression, surrounding objects, or the broader scene context.

Modern image editors combine textual instructions with attention and spatial controls~\citep{meng2022sdedit,hertz2023prompt,brooks2023instructpix2pix,avrahami2022blended,zhang2023controlnet,zhang2023magicbrush}. Such controls are effective when the desired attribute or region is explicit, but they leave the source-side decision implicit for affective editing~\citep{fu2024guiding,chen2026face}. Existing affective editors expose a related tension: methods that produce stronger emotion shifts can introduce artifacts or structural drift, whereas more constrained methods preserve the source but make weak or misplaced changes when the relevant cues are unspecified~\citep{yang2025emoedit,ye2025moodifier,yin2026mood}. The ambiguity is amplified in complex scenes, where affect is distributed across multiple objects and contextual relations rather than concentrated in a face or dominant subject.

Such distributed affect makes source-emotion evidence difficult to identify and use for editing. Visual emotion datasets and explanatory annotations characterize content associated with perceived affect~\citep{peng2015mixedbag,kosti2020emotic,achlioptas2021artemis,yang2023emoset}. Complementary attribution methods expose model evidence through gradients, local surrogates, or perturbations~\citep{selvaraju2017gradcam,sundararajan2017integrated,ribeiro2016lime,petsiuk2018rise,fong2017meaningful,goyal2019counterfactual}. These resources support recognition and diagnosis, but they do not directly yield reusable, signed object-level contributions that distinguish evidence supporting the source emotion from evidence opposing it. Moreover, localization alone does not specify how to introduce the target emotion while preserving unrelated content~\citep{yang2024cross,ye2024dual,hong2026emostyle}. However, effective AIE consistently requires two coupled decisions: identify and characterize the source evidence, then translate that evidence and the target emotion into an appropriate editing operation. Neither decision follows from the target label alone.

To address this need, we introduce \textbf{IECLA}, which learns signed, object-level source-emotion contributions from counterfactual interventions and exposes them as an editing interface. At inference, a frozen evaluator predicts the source emotion from the input, and the contribution predictor assigns regional scores without additional annotations. A target-aware agent combines these signals with the desired emotion to construct constrained instructions for existing editors.

In summary, our main contributions are summarized as follows:
\begin{itemize}
    \item We formulate signed object-level counterfactual contributions as a supervision signal for source-emotion attribution. The learned contributions distinguish editable source-emotion evidence from regions to preserve, without requiring a source-emotion label at inference.
    \item We build a unified attribution-guided framework that connects the learned evidence to target-aware image editing. It produces interpretable emotion heatmaps and converts attributed evidence into guidance for image editors without retraining their parameters.
    \item We validate IECLA through multiple benchmarks, experiments, and backbones. The results show more faithful attribution and stronger target-emotion expression while preserving source structure and naturalness.
\end{itemize}

\section{Related Work}
\label{sec:related-work}

\paragraph{Visual emotion understanding and attribution.}
Visual affective computing models depict and elicit viewer's emotions with categorical labels, response distributions, scene context, and textual explanations~\citep{peng2015mixedbag,kosti2020emotic,achlioptas2021artemis,yang2023emoset}. Vision and language approaches further organize affective knowledge into interpretable rationales~\citep{man2025vaeer}. Complementary attribution methods expose model evidence through gradients~\citep{selvaraju2017gradcam,jiang2021layercam,sundararajan2017integrated}, local surrogates~\citep{ribeiro2016lime}, or perturbations~\citep{petsiuk2018rise,fong2017meaningful}; EmoCAM combines class-activation maps with detected objects for corpus-scale emotion analysis~\citep{doulfoukar2024emocam}, and counterfactual explanations identify evidence whose alteration changes a prediction~\citep{goyal2019counterfactual}. These resources support recognition and diagnosis, but their supervision or outputs are not defined as reusable, signed object-level controls for affective editing. Perturbation-based estimates require multiple evaluator queries, whereas visually plausible maps do not need to be strictly faithful~\citep{adebayo2018sanity,zhang2026creatidesign,zhang2025creatiposter}. IECLA builds on this line by learning intervention-grounded contributions with an explicit sign and an editor-facing interface.
\vspace{-5pt}

\paragraph{Affective image editing and agentic control.}
General editors use partial noising~\citep{meng2022sdedit}, spatial and attention controls~\citep{avrahami2022blended,zhang2023controlnet,hertz2023prompt}, or learned instruction following~\citep{brooks2023instructpix2pix}. Affective editing additionally requires translating emotion into image-dependent changes~\citep{yuan2026coemogen, wei2025omniedit}. EmoEdit learns an emotion adapter from content-aware edit pairs~\citep{yang2025emoedit}, Moodifier derives contextual attributes through affect-specialized representations and multimodal reasoning~\citep{ye2025moodifier}, and MooD provides continuous valence--arousal control~\citep{yin2026mood}. EmoAgent decomposes editing into planning, execution, and critique~\citep{mao2025emoagent}, while EmoKGEdit combines affective localization, knowledge-guided cue transformation, and structure-aware rendering~\citep{zhang2026emokgedit,wang2026multi}. These approaches advance target-side control. IECLA instead learns signed source-emotion contributions as an explicit interface to target-aware editing.
\vspace{-5pt}

\section{Method}
\label{sec:method}

\subsection{Task and Framework Overview}

Given a source image $I$ and a target emotion $t$, affective image editing seeks to evoke $t$ while preserving unrelated content. IECLA infers the source emotion $y$ from $I$, estimates how candidate regions support or oppose $y$, and uses this signed evidence to guide an existing editor.

\begin{figure*}[t]
    \centering
    \setlength{\abovecaptionskip}{-5pt}
    \setlength{\belowcaptionskip}{0pt}
    \includegraphics[width=\textwidth,pagebox=cropbox]{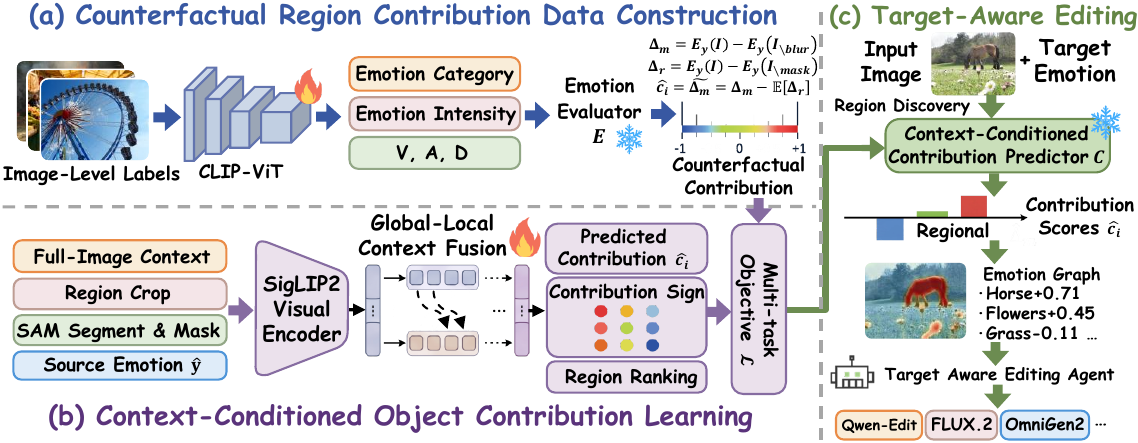}
    \caption{IECLA framework. (a) Counterfactual interventions construct signed supervision. (b) Context-conditioned learning predicts regional contributions. (c) Signed predictions form emotion heatmaps and ground target-aware editing operations.}
    \label{fig:overview}
    \vspace{-10pt}
\end{figure*}

\Cref{fig:overview} shows the complete pipeline. During training, counterfactual interventions construct region-level contributions, and a context-conditioned predictor learns to estimate them from the original image and masks. At inference, the predicted contributions form an emotion heatmap and separate editable from preservation evidence. A target-aware agent combines this evidence with $t$, verifies feasible changes, and selects a budgeted editing plan. After structural review and instruction auditing, the compiled instruction is passed to the existing editor.

\subsection{Counterfactual Contribution Label Construction}
\label{sec:counterfactual-labels}

To obtain signed region-level supervision, we first infer the source emotion, then measure how individual regions affect its score. Image-level supervision identifies the perceived affect but does not reveal which regions produce it or how they contribute. We use evaluator $E$, trained with EmoVerse \citep{guo2025emoverse} image-level supervision, to infer the source emotion from the full image:
\begin{equation}
y=\arg\max_{e\in\mathcal{E}} E_e(I),
\label{eq:source-emotion-inference}
\end{equation}
where $\mathcal{E}$ is the emotion vocabulary and $E_e(I)$ is the probability assigned to emotion $e$. At inference, the same frozen evaluator is evaluated only on the unmodified image to obtain $y$; the user specifies the target emotion $t$, and no source-emotion annotation is required.

To assess each candidate region relative to a comparable blur baseline, we compare its blur intervention against a matched reference-mask intervention. Grounding DINO~\citep{liu2024groundingdino} and SAM~\citep{kirillov2023sam} provide candidate regions indexed by $i$. Let $I_{i,\setminus\mathrm{blur}}$ denote the intervention that blurs candidate region $i$, and let $I_{i,\setminus\mathrm{mask}}$ denote its matched reference-mask intervention. Their effects on the source-emotion response are defined as:
\begin{equation}
\begin{aligned}
\Delta_m = E_y(I)-E_y(I_{i,\setminus\mathrm{blur}}),~\Delta_r= E_y(I)-E_y(I_{i,\setminus\mathrm{mask}}),~c_i = \Delta_m-\mathbb{E}[\Delta_r].
\end{aligned}
\label{eq:counterfactual-contribution}
\end{equation}
The correction estimates the candidate's effect relative to the average response to comparable interventions. Here, $E_y(\cdot)$ denotes the frozen evaluator response for the inferred source emotion $y$, whereas $\mathbb{E}[\Delta_r]$ averages the effects of area- and position-matched reference interventions. We retain $\Delta_m$ because averaging it would remove the region-specific signal that $c_i$ is intended to preserve. The resulting corrected contribution $c_i$ is the training target: positive and negative values denote effects above and below the reference baseline, respectively.

The resulting $c_i$ provides the direct supervision interface for the predictor in Section~\ref{sec:contribution-prediction}. Repeating the intervention procedure over candidate regions yields one signed training instance per region.

\subsection{Context-Conditioned Signed Contribution Prediction}
\label{sec:contribution-prediction}

To estimate $c_i$ without interventions, we train a context-conditioned predictor that operates on a single unmodified image. Built on the SigLIP2 vision backbone~\citep{tschannen2025siglip2}, the predictor integrates region-conditioned visual features, the inferred source emotion, and geometric metadata to recover the signed contribution target constructed in Section~\ref{sec:counterfactual-labels}.

We model the contextual dependence of regional affect by constructing three visual views at both global and local scales: the candidate region, its complementary context, and the complete image. At each scale $s\in\{g,\ell\}$, the backbone encodes these views as $\mathbf{r}_i^s$, $\mathbf{k}_i^s$, and $\mathbf{a}^s$, respectively. We fuse the three representations with candidate--context difference and product features:
\begin{equation}
\begin{aligned}
\mathbf{v}_i^s={}&[\mathbf{r}_i^s;\mathbf{k}_i^s;\mathbf{a}^s;
\mathbf{r}_i^s-\mathbf{k}_i^s;\mathbf{r}_i^s\odot\mathbf{k}_i^s],\\[-1pt]
\mathbf{h}_i={}&[\mathbf{v}_i^g;\mathbf{v}_i^\ell;\mathbf{e}_i;
\operatorname{Emb}(y);\mathbf{u}_i;\mathbf{g}_i].
\end{aligned}
\label{eq:multiscale-features}
\end{equation}
Here, $\mathbf{e}_i$ encodes region semantics, $\mathbf{u}_i$ summarizes the evaluator state before intervention, and $\mathbf{g}_i$ contains geometric and quality metadata. From the resulting representation $\mathbf{h}_i$, the prediction head estimates a nonnegative contribution magnitude and probabilities over negative, neutral, and positive signs, which are combined into a signed prediction:
\begin{equation}
\widehat{c}_i=m_i\left(p_i^{+}-p_i^{-}\right),
\label{eq:signed-output}
\end{equation}
where $m_i=\operatorname{softplus}(z_i)$, and $p_i^{+}$ and $p_i^{-}$ are the positive and negative probabilities; the neutral class contributes zero. The predictor must recover not only the value of $c_i$, but also its direction and the relative importance of regions within the same image. We therefore optimize
\begin{equation}
\begin{aligned}
\mathcal{L}={}&\mathcal{L}_{\mathrm{Huber}}^{w}
 +0.25\mathcal{L}_{\mathrm{abs}}
 +0.10\mathcal{L}_{\mathrm{signed}}+0.20\mathcal{L}_{\mathrm{focal}}
 +0.10\mathcal{L}_{\mathrm{aux}}.
\end{aligned}
\label{eq:training-objective}
\end{equation}
The training objective jointly constrains contribution magnitude, within-image ranking, sign, and affective auxiliaries. The weighted Huber term learns contribution magnitude, the absolute-ranking term preserves regional importance, and the signed-ranking and focal terms supervise direction. The auxiliary term retains emotion-intensity and VAD information. Appendix~\ref{app:implementation-details} provides loss definitions and weighting details. Once trained, the predictor directly yields signed regional evidence without counterfactual interventions.

\subsection{Attribution-Grounded Target Editing}

Upon completing training, we apply the predictor to candidate regions identified by Region Discovery in the source image $I$. The resulting signed scores $\widehat{c}_i$ are mapped onto their regions to form an interpretable attribution heatmap, showing which content supports or opposes the inferred source emotion $y$. These predictions provide the attribution grounding for subsequent target editing.

The target-aware agent translates the signed regional evidence and target emotion $t$ into an executable, image-specific editing plan. It verifies the attribution against the visible content, identifies the primary emotional carrier, and selects the least disruptive operations. The primary local action is anchored to attributed source-emotion evidence, while optional atmosphere or minor-object actions are included only when the scene supports; negative evidence imposes preservation constraints.

Finally, a visual and structural critic checks the plan for executability and content preservation. A deterministic compiler assembles the approved operations into an editing prompt, and an integrated-instruction auditor checks it for conflicts before it is passed to the editor. Detailed routing and validation procedures are provided in Appendices~\ref{app:agent-template} and~\ref{app:implementation-details}.

\begin{figure}[!t]
\centering
\includegraphics[width=\columnwidth,pagebox=cropbox]{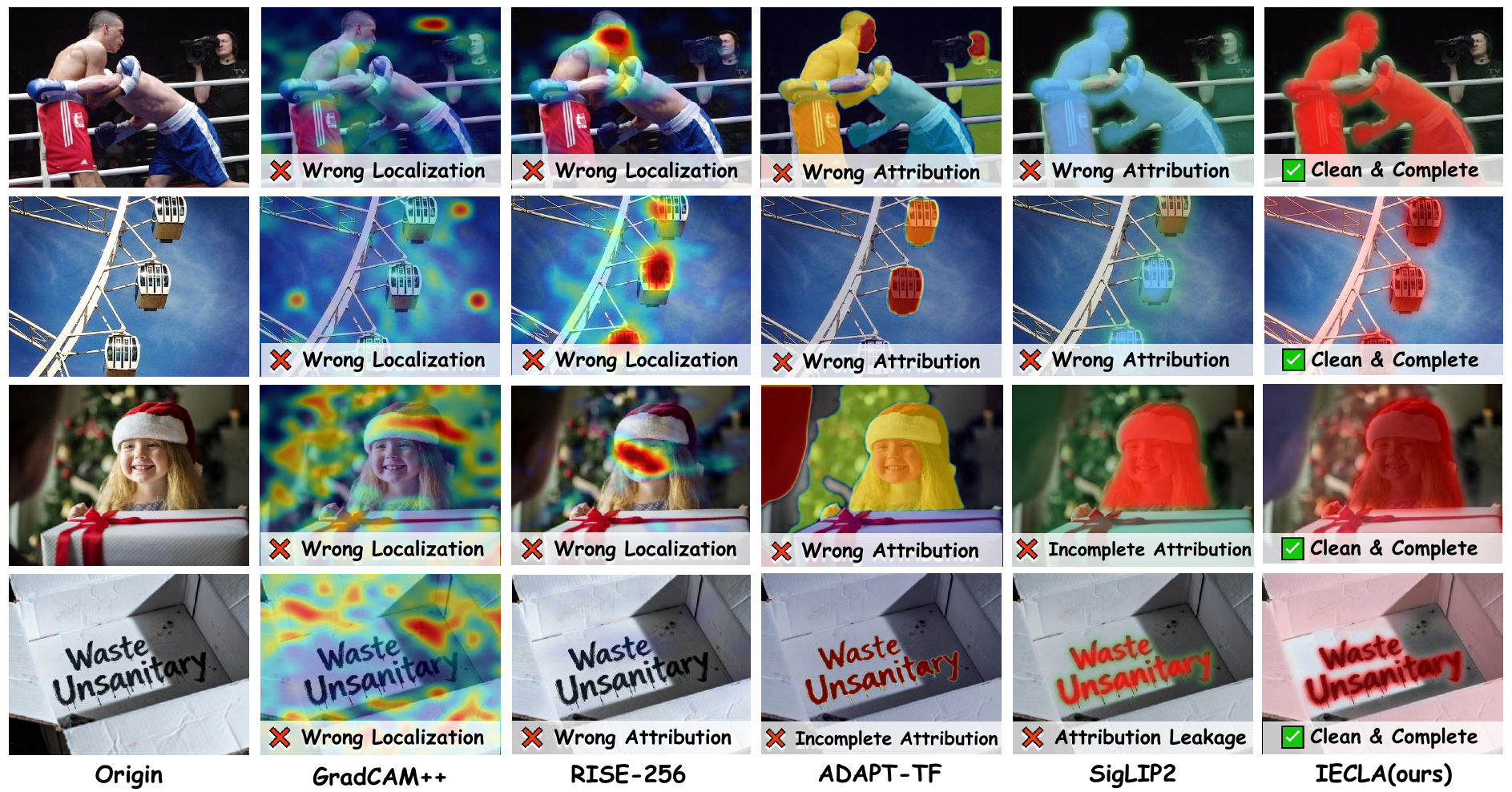}
\vspace{-20pt}
\caption{Representative emotion attribution maps on diverse scenes. Existing approaches mislocalize affective evidence or omit opposing regions. IECLA produces more accurate heatmaps.}
\vspace{-10pt}
\label{fig:heatmap}
\end{figure}

\section{Experiments}
\label{sec:experiments}

\subsection{Experimental Setup}
\label{sec:experimental-setup}

The locked attribution test contains 7,278 images and 53,180 SAM masks. The training, validation, and test subsets are strictly disjoint. Training uses three NVIDIA A800 GPUs. Cross-dataset evaluation excludes the external datasets' original labels from training and teacher-emotion selection. Training details and the eight editing emotions are provided in Appendix~\ref{app:implementation-details}.

Main editing comparisons and ablations cover 680 source images paired with all eight target emotions, yielding 5,440 tasks across Qwen-Edit~\citep{wu2025qwenimage}, InstructPix2Pix~\citep{brooks2023instructpix2pix}, FLUX.2~\citep{blackforestlabs2026flux2}, LongCat~\citep{longcat2025image}, OmniGen2~\citep{wu2025omnigen2}, and HiDream~\citep{cai2025hidreami1}; exact editor variants are listed in Appendix~\ref{app:implementation-details}. We compare each editor with its native pipeline using matched source images, targets, seeds, samplers, and resolutions. The teacher \(E\) provides source-emotion inference and training supervision, while the contribution predictor and agent perform candidate scoring and editing planning. Target Success and \(\Delta p\) measure teacher-aligned responses under \(E\), whereas Emo-A and Emo-S provide a second classifier view; none of these post-hoc scores is used for planning. Attribution ranking and localization metrics are computed per image and averaged; Sign Macro-F1 and continuous contribution errors are computed over all valid candidate regions. Confidence intervals use image-level cluster bootstrap.

\subsection{Attribution Benchmark Results}
\label{sec:attribution-evaluation}

To evaluate the accuracy of IECLA's source-emotion attribution, we compare it with two baseline families: conventional heatmap attribution methods and signed attribution models built on CLIP~\citep{radford2021clip}, DINOV3~\citep{simeoni2025dinov3}, and SigLIP2~\citep{tschannen2025siglip2}. The signed models are trained on the same counterfactual contribution targets, whereas the nonnegative heatmap methods are evaluated only on unsigned ranking metrics. Table~\ref{tab:teacher-baselines} reports region ranking, contribution direction, and magnitude against the frozen evaluator's counterfactual contributions; this benchmark measures teacher-defined fidelity rather than human causal attribution.

\begin{figure}[!t]
\centering
\includegraphics[width=\columnwidth,pagebox=cropbox]{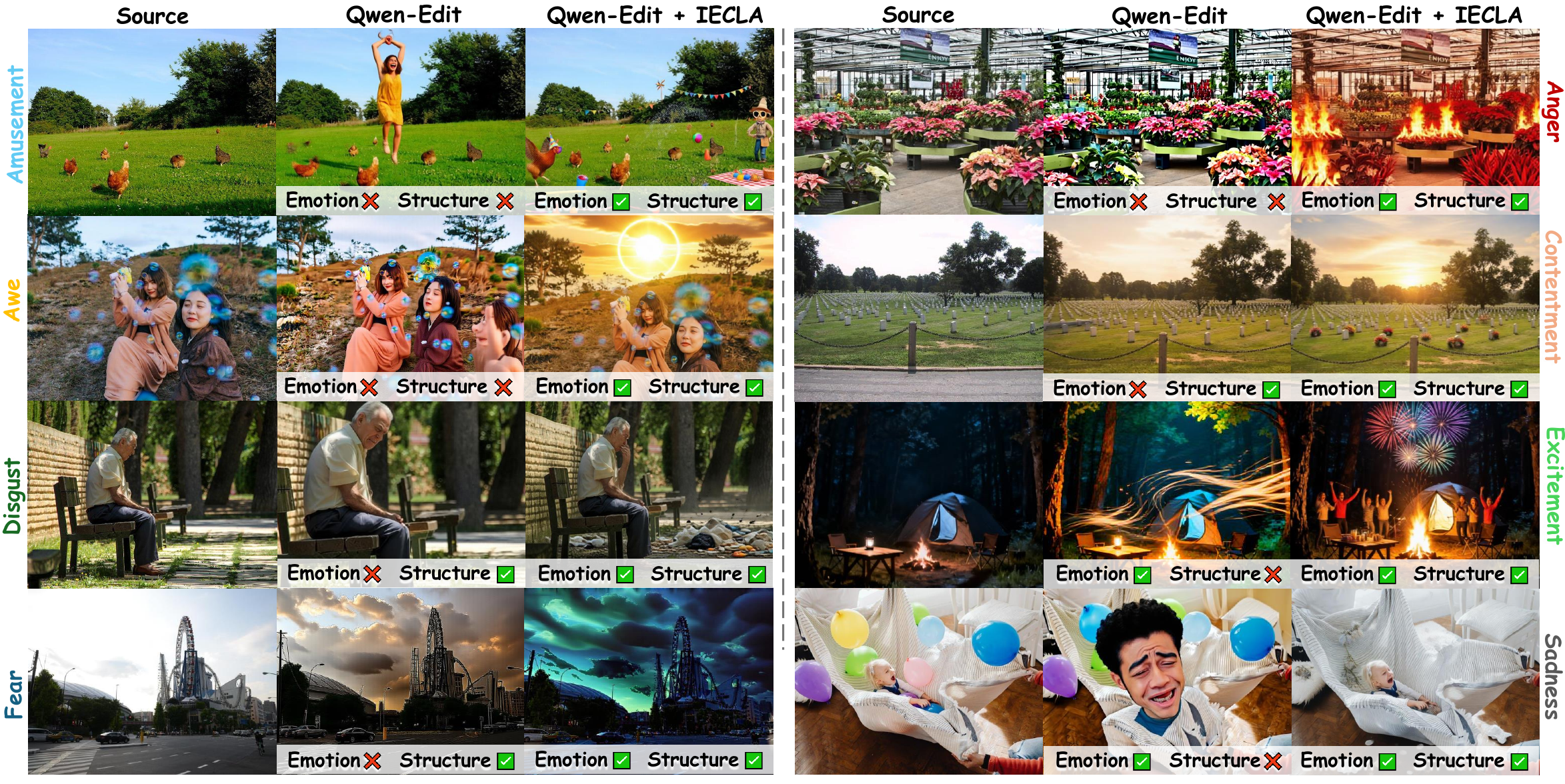}
\vspace{-14pt}
\caption{Eight-emotion editing with Qwen-Edit, with and without IECLA. IECLA consistently elicits the requested emotion while preserving scene structure across all eight target categories.}
\vspace{-8pt}
\label{fig:editing-examples}
\end{figure}

\begin{table}[!t]
\begingroup
\centering
\setlength{\tabcolsep}{4pt}
\renewcommand{\arraystretch}{0.90}
\setlength{\abovecaptionskip}{2pt}
\setlength{\belowcaptionskip}{3pt}
\caption{Comparison with attribution baselines on the locked test set. IECLA achieves the strongest performance across ranking, sign agreement, and contribution regression.}
\label{tab:teacher-baselines}
\normalsize
\makebox[\columnwidth][c]{%
\scalebox{0.762}{%
\begin{tabular}{lcccccccc}
\toprule
Method & Imp. $\rho\uparrow$ & Signed $\rho\uparrow$ & Top-1$\uparrow$ & NDCG@5$\uparrow$ & Macro-F1$\uparrow$ & MAE$\downarrow$ & RMSE$\downarrow$ & $R^2\uparrow$ \\
\midrule
\rowcolor{gray!15}\multicolumn{9}{l}{\textit{Heatmap attribution methods}} \\
Ablation-CAM~\citeyearpar{desai2020ablationcam} & -0.0452 & -- & 0.2279 & 0.5877 & -- & -- & -- & -- \\
Grad-CAM~\citeyearpar{selvaraju2017gradcam} & 0.0328 & -- & 0.2525 & 0.6313 & -- & -- & -- & -- \\
Direct attention & 0.0432 & -- & 0.2376 & 0.6321 & -- & -- & -- & -- \\
Grad-CAM++~\citeyearpar{chattopadhay2018gradcampp} & 0.0690 & -- & 0.2567 & 0.6443 & -- & -- & -- & -- \\
RISE-256~\citeyearpar{petsiuk2018rise} & 0.0861 & -- & 0.2521 & 0.6428 & -- & -- & -- & -- \\
ADAPT-TF~\citeyearpar{yao2026adapt} & 0.1214 & -- & 0.2700 & 0.6592 & -- & -- & -- & -- \\
EigenCAM~\citeyearpar{muhammad2020eigencam} & 0.1249 & -- & 0.2751 & 0.6705 & -- & -- & -- & -- \\
HiResCAM~\citeyearpar{draelos2020use} & 0.1986 & -- & 0.2987 & 0.7051 & -- & -- & -- & -- \\
LayerCAM~\citeyearpar{jiang2021layercam} & 0.2176 & -- & 0.2991 & 0.7078 & -- & -- & -- & -- \\
\midrule
\rowcolor{gray!15}\multicolumn{9}{l}{\textit{Vision-encoder contribution methods}} \\
CLIP~\citeyearpar{radford2021clip} & 0.3534 & 0.1715 & 0.4325 & 0.7918 & 0.3273 & 0.1145 & 0.2009 & 0.2036 \\
DINOV3~\citeyearpar{simeoni2025dinov3} & 0.3576 & 0.1692 & 0.4376 & 0.7940 & 0.3387 & 0.1116 & 0.2081 & 0.1458 \\
SigLIP2~\citeyearpar{tschannen2025siglip2} & \underline{0.3819} & \underline{0.2314} & \underline{0.4445} & \underline{0.8050} & \underline{0.3454} & \underline{0.1020} & \underline{0.1833} & \underline{0.3370} \\
\rowcolor{orange!25!yellow!30}\textbf{IECLA (ours)} & \textbf{0.5885} & \textbf{0.5088} & \textbf{0.5808} & \textbf{0.8915} & \textbf{0.3703} & \textbf{0.0939} & \textbf{0.1743} & \textbf{0.4005} \\
\bottomrule
\vspace{-30pt}
\end{tabular}%
}%
}
\endgroup
\end{table}

Two patterns distinguish contribution learning from generic saliency. The heatmap methods recover some relevant regions, but their nonnegative outputs cannot represent evidence that opposes the source emotion. The learned vision-encoder models improve regional ranking, while IECLA further leads them in ranking, signed agreement, and regression. This joint improvement is important because the teacher scores concentrate near zero: a near-neutral predictor can attain a low MAE without recovering candidate order or direction. IECLA instead captures which regions matter, whether they support or oppose the source emotion, and their relative contribution strength, Figure~\ref{fig:heatmap} illustrates the corresponding localization in complex scenes.

\subsection{Comparison with Native Editors}
\label{sec:editing-baselines}

\noindent

\vspace{0pt}
To assess transfer across editing backbones, Table~\ref{tab:editing-native} compares IECLA-guided editing with native emotion-request pipelines on the 5,440 matched tasks defined in Section~\ref{sec:experimental-setup}.

Native editors exhibit two complementary failure modes: conservative edits often preserve the source but under-express the requested emotion, whereas stronger changes can alter subject identity or surrounding structure. Table~\ref{tab:editing-native} shows that IECLA shifts this trade-off rather than merely increasing edit strength. Across all six editors, affective outcomes improve together with higher PSNR and SSIM and lower LPIPS. The common direction across editors with different native behavior supports attribution and target-aware planning as a transferable control interface, while the variation in gain magnitude reflects the execution capacity of each backbone. Appendix~\ref{app:stability} examines whether the gains persist across seeds and resolutions.

\begin{figure}[!t]
\centering
\includegraphics[width=\linewidth,pagebox=cropbox]{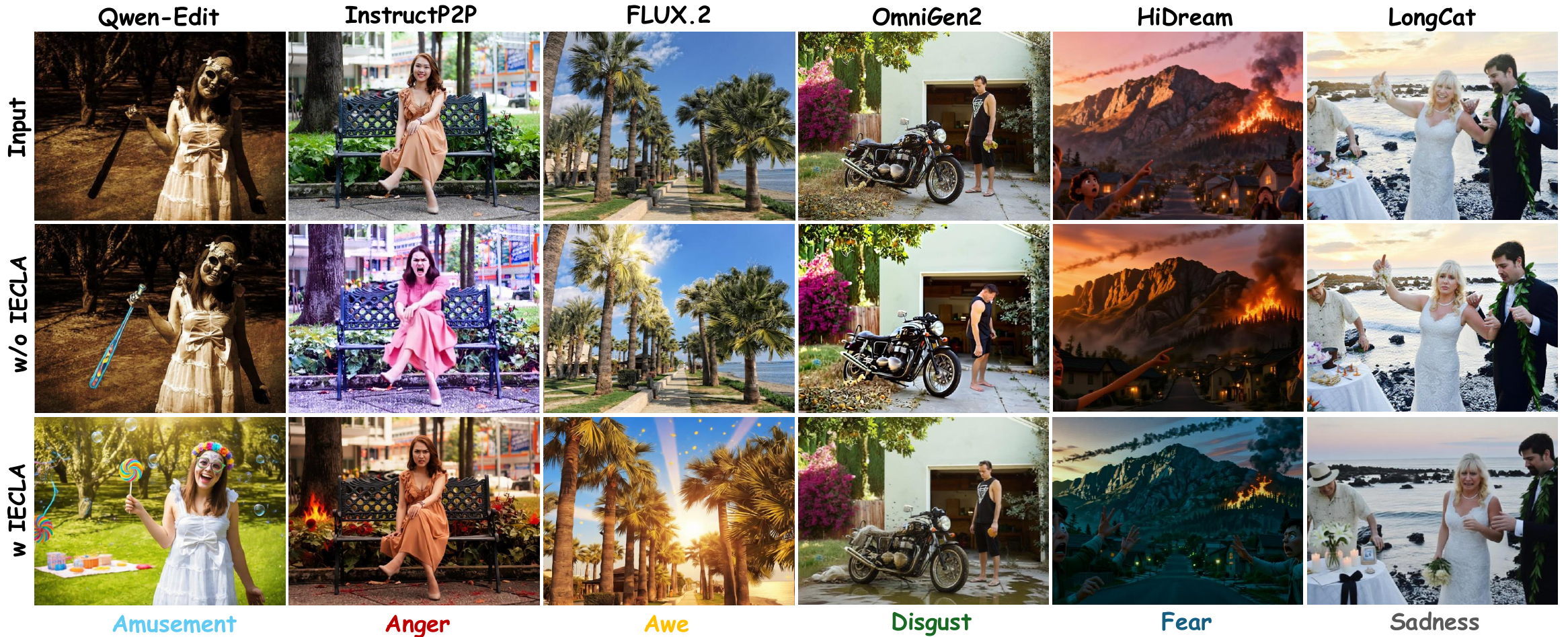}
\vspace{-18pt}
\caption{Qualitative comparison across six editing backbones, with and without IECLA. IECLA improves target-emotion expression across diverse editors while preserving source structure.}
\vspace{-8pt}
\label{fig:editing-qualitative}
\end{figure}

Figures~\ref{fig:teaser},~\ref{fig:editing-examples}, and~\ref{fig:editing-qualitative} illustrate the same trend in multi-object and context-dependent scenes: IECLA directs edits toward attributed cues while constraining unrelated content, avoiding the subject drift or scene-wide changes visible in several native outputs. Appendix~\ref{app:additional-editing-results} provides further Qwen-Edit examples and analyzes the corresponding failure modes in Figures~\ref{fig:additional-editing-results} and~\ref{fig:additional-native-comparison}.

\begin{table}[!t]
\begingroup
\centering
\setlength{\tabcolsep}{2pt}
\renewcommand{\arraystretch}{0.92}
\setlength{\abovecaptionskip}{0pt}
\setlength{\belowcaptionskip}{2pt}
\newcommand{\reldelta}[1]{\cellcolor{orange!25!yellow!30}\textbf{#1}}
\caption{Comparison between native and IECLA-guided editing across six backbones. IECLA consistently improves target-emotion expression and structural preservation.}
\label{tab:editing-native}
\normalsize
\makebox[\linewidth][c]{%
\scalebox{0.762}{%
\begin{tabular}{lccrccrccr}
\toprule

\rowcolor{gray!15}\textit{Affective outcomes}
& \multicolumn{3}{c}{Target Success$\uparrow$}
& \multicolumn{3}{c}{Emo-A$\uparrow$}
& \multicolumn{3}{c}{Emo-S$\uparrow$} \\

\cmidrule(lr){2-4} \cmidrule(lr){5-7} \cmidrule(lr){8-10}

Editor 
& Native & IECLA-guided & Rel. $\Delta$
& Native & IECLA-guided & Rel. $\Delta$
& Native & IECLA-guided & Rel. $\Delta$ \\

\midrule

Qwen-Edit~\citeyearpar{wu2025qwenimage}
& 0.2996 & 0.3579 & \reldelta{+19.5\%}
& 0.2588 & 0.3081 & \reldelta{+19.0\%}
& 0.1843 & 0.2173 & \reldelta{+17.9\%} \\

InstructPix2Pix~\citeyearpar{brooks2023instructpix2pix}
& 0.1906 & 0.2127 & \reldelta{+11.6\%}
& 0.2329 & 0.2704 & \reldelta{+16.1\%}
& 0.1103 & 0.1508 & \reldelta{+36.7\%} \\

FLUX.2~\citeyearpar{blackforestlabs2026flux2}
& 0.2283 & 0.2643 & \reldelta{+15.8\%}
& 0.1465 & 0.1522 & \reldelta{+3.9\%}
& 0.0867 & 0.1508 & \reldelta{+73.9\%} \\

LongCat~\citeyearpar{longcat2025image}
& 0.2774 & 0.3099 & \reldelta{+11.7\%}
& 0.1801 & 0.1844 & \reldelta{+2.4\%}
& 0.0976 & 0.1053 & \reldelta{+7.9\%} \\

OmniGen2~\citeyearpar{wu2025omnigen2}
& 0.2752 & 0.3182 & \reldelta{+15.6\%}
& 0.2285 & 0.2877 & \reldelta{+25.9\%}
& 0.1229 & 0.1811 & \reldelta{+47.4\%} \\

HiDream~\citeyearpar{cai2025hidreami1}
& 0.2634 & 0.2947 & \reldelta{+11.8\%}
& 0.2259 & 0.2461 & \reldelta{+8.9\%}
& 0.0735 & 0.1130 & \reldelta{+53.7\%} \\

\midrule

\rowcolor{gray!15}\textit{Structural preservation}
& \multicolumn{3}{c}{PSNR$\uparrow$}
& \multicolumn{3}{c}{SSIM$\uparrow$}
& \multicolumn{3}{c}{LPIPS$\downarrow$} \\

\cmidrule(lr){2-4} \cmidrule(lr){5-7} \cmidrule(lr){8-10}

Editor 
& Native & IECLA-guided & Rel. $\Delta$
& Native & IECLA-guided & Rel. $\Delta$
& Native & IECLA-guided & Rel. $\Delta$ \\

\midrule

Qwen-Edit~\citeyearpar{wu2025qwenimage}
& 15.4985 & 20.3046 & \reldelta{+31.0\%}
& 0.5790 & 0.6137 & \reldelta{+6.0\%}
& 0.3698 & 0.3440 & \reldelta{-7.0\%} \\

InstructPix2Pix~\citeyearpar{brooks2023instructpix2pix}
& 23.4641 & 27.7156 & \reldelta{+18.1\%}
& 0.6017 & 0.6682 & \reldelta{+11.1\%}
& 0.3890 & 0.1840 & \reldelta{-52.7\%} \\

FLUX.2~\citeyearpar{blackforestlabs2026flux2}
& 30.8471 & 31.9981 & \reldelta{+3.7\%}
& 0.4166 & 0.4346 & \reldelta{+4.3\%}
& 0.4190 & 0.3249 & \reldelta{-22.5\%} \\

LongCat~\citeyearpar{longcat2025image}
& 16.0436 & 16.6780 & \reldelta{+4.0\%}
& 0.4555 & 0.4876 & \reldelta{+7.0\%}
& 0.3224 & 0.2923 & \reldelta{-9.3\%} \\

OmniGen2~\citeyearpar{wu2025omnigen2}
& 14.3197 & 17.4542 & \reldelta{+21.9\%}
& 0.4854 & 0.5966 & \reldelta{+22.9\%}
& 0.2881 & 0.1820 & \reldelta{-36.8\%} \\

HiDream~\citeyearpar{cai2025hidreami1}
& 14.7940 & 19.9107 & \reldelta{+34.6\%}
& 0.4435 & 0.5366 & \reldelta{+21.0\%}
& 0.2893 & 0.1758 & \reldelta{-39.2\%} \\

\bottomrule
\vspace{-20pt}
\end{tabular}%
}%
}
\endgroup
\end{table}

\subsection{Cross-Dataset Evaluation}
\label{sec:cross-dataset}

To test transfer beyond the training distribution, we evaluate IECLA on 3,370 images from five external datasets: COCO~\citep{lin2014coco}, EmoSet~\citep{yang2023emoset}, FI~\citep{you2016building}, Emotion6~\citep{peng2015mixedbag}, and EmoArt~\citep{zhang2025emoart}. We then test a downstream editing task on Qwen-Edit, comparing its native pipeline with IECLA-guided editing.

\begin{table}[!t]
\begingroup
\centering
\setlength{\tabcolsep}{2pt}
\renewcommand{\arraystretch}{0.92}
\setlength{\abovecaptionskip}{2pt}
\setlength{\belowcaptionskip}{3pt}
\newcommand{\crossdelta}[1]{\cellcolor{orange!25!yellow!30}\textbf{#1}}
\caption{Cross-dataset editing with Qwen-Edit. IECLA improves affective expression and structural preservation across all five datasets.}
\label{tab:cross-dataset}
\normalsize
\makebox[\linewidth][c]{%
\scalebox{0.762}{%
\begin{tabular}{@{}lccc ccc ccc@{}}
\toprule
\rowcolor{gray!15}\textit{Affective outcomes}
& \multicolumn{3}{c}{Target Success $\uparrow$}
& \multicolumn{3}{c}{Emo-A $\uparrow$}
& \multicolumn{3}{c}{Emo-S $\uparrow$} \\
\cmidrule(lr){2-4} \cmidrule(lr){5-7} \cmidrule(lr){8-10}
Dataset & Native & IECLA-guided & Rel. $\Delta$
& Native & IECLA-guided & Rel. $\Delta$
& Native & IECLA-guided & Rel. $\Delta$ \\
\midrule
COCO
& 0.2188 & 0.2939 & \crossdelta{+34.3\%}
& 0.1687 & 0.2375 & \crossdelta{+40.8\%}
& 0.0394 & 0.1015 & \crossdelta{+157.6\%} \\
EmoArt
& 0.2757 & 0.3664 & \crossdelta{+32.9\%}
& 0.1627 & 0.2397 & \crossdelta{+47.3\%}
& 0.0425 & 0.1120 & \crossdelta{+163.5\%} \\
EmoSet
& 0.2642 & 0.3396 & \crossdelta{+28.5\%}
& 0.1949 & 0.2642 & \crossdelta{+35.6\%}
& 0.0685 & 0.1114 & \crossdelta{+62.6\%} \\
Emotion6
& 0.1527 & 0.2611 & \crossdelta{+71.0\%}
& 0.1587 & 0.2168 & \crossdelta{+36.6\%}
& 0.0313 & 0.1065 & \crossdelta{+240.3\%} \\
FI
& 0.2704 & 0.3082 & \crossdelta{+14.0\%}
& 0.1761 & 0.2390 & \crossdelta{+35.7\%}
& 0.0593 & 0.0937 & \crossdelta{+58.0\%} \\
\midrule
\rowcolor{gray!15}\textit{Structural preservation}
& \multicolumn{3}{c}{PSNR $\uparrow$}
& \multicolumn{3}{c}{SSIM $\uparrow$}
& \multicolumn{3}{c}{LPIPS $\downarrow$} \\
\cmidrule(lr){2-4} \cmidrule(lr){5-7} \cmidrule(lr){8-10}
Dataset & Native & IECLA-guided & Rel. $\Delta$
& Native & IECLA-guided & Rel. $\Delta$
& Native & IECLA-guided & Rel. $\Delta$ \\
\midrule
COCO
& 12.8522 & 16.3243 & \crossdelta{+27.0\%}
& 0.2761 & 0.3918 & \crossdelta{+41.9\%}
& 0.4932 & 0.2427 & \crossdelta{-50.8\%} \\
EmoArt
& 14.4890 & 15.7233 & \crossdelta{+8.5\%}
& 0.3429 & 0.3659 & \crossdelta{+6.7\%}
& 0.2339 & 0.2219 & \crossdelta{-5.1\%} \\
EmoSet
& 12.7428 & 17.0118 & \crossdelta{+33.5\%}
& 0.3237 & 0.4595 & \crossdelta{+42.0\%}
& 0.4201 & 0.2181 & \crossdelta{-48.1\%} \\
Emotion6
& 13.5218 & 17.4566 & \crossdelta{+29.1\%}
& 0.3944 & 0.5366 & \crossdelta{+36.1\%}
& 0.3534 & 0.1719 & \crossdelta{-51.4\%} \\
FI
& 14.2610 & 17.6349 & \crossdelta{+23.7\%}
& 0.4066 & 0.5093 & \crossdelta{+25.3\%}
& 0.3389 & 0.1675 & \crossdelta{-50.6\%} \\
\bottomrule
\vspace{-20pt}
\end{tabular}%
}%
}
\endgroup
\end{table}

Across the five external datasets, IECLA remains the leading attribution predictor (Appendix~\ref{app:cross-dataset-details}) and improves downstream editing across natural scenes, affective photographs, and artwork (Table~\ref{tab:cross-dataset}). All three preservation metrics improve on every dataset. EmoArt shows smaller structural gains, consistent with artistic and abstract images distributing affective cues across palette, texture, and composition rather than a localized region. Even in this domain, the positive gains show that stronger target-emotion expression does not sacrifice structural preservation.

\vspace{-3pt}
\subsection{Ablation Study}
\label{sec:system-ablation}
\vspace{-3pt}

\begin{wrapfigure}{R}{0.61\textwidth}
\vspace{-10pt}
\centering
\includegraphics[width=\linewidth,pagebox=cropbox]{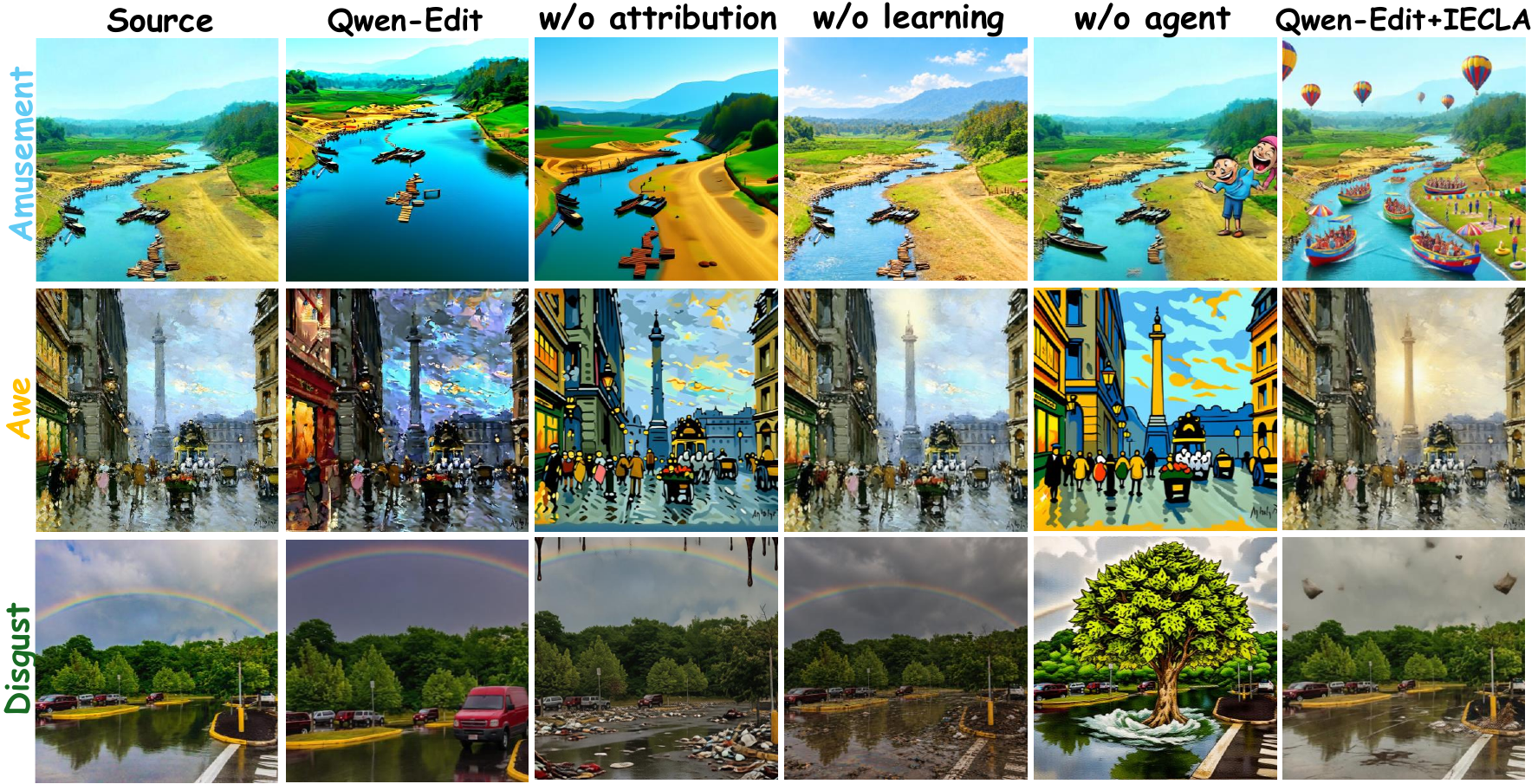}
\vspace{-15pt}
\caption{Qualitative ablation on Qwen-Image-Edit. IECLA better expresses target emotions while preserving structure than the baseline and component-removed variants.}
\label{fig:ablation-qualitative}
\vspace{-5pt}
\end{wrapfigure}

Table~\ref{tab:system-ablation} reports systematic ablations of each core component of the IECLA framework to assess its contribution, with qualitative examples in Figure~\ref{fig:ablation-qualitative}. Appendix~\ref{app:loss-ablation} provides additional ablation of editing controls and loss ablations, with the latter summarized in Figure~\ref{fig:reference-validation}(b).

\begin{table}[t]
\begingroup
\centering
\setlength{\tabcolsep}{2.3pt}
\renewcommand{\arraystretch}{0.94}
\setlength{\abovecaptionskip}{2pt}
\setlength{\belowcaptionskip}{3pt}
\caption{Ablation of IECLA's attribution, learning, and agent modules.}
\label{tab:system-ablation}
\resizebox{\textwidth}{!}{%
\begin{tabular}{@{}lcccc@{\hspace{9pt}}cccccc@{}}
\toprule
\raisebox{-1.6ex}[0pt][0pt]{Condition} &
\multicolumn{4}{c}{\cellcolor{gray!15}\textit{Target-emotion outcomes}} &
\multicolumn{6}{c}{\cellcolor{gray!15}\textit{Structural preservation}} \\
\cmidrule(lr){2-5}\cmidrule(l){6-11}
& Success $\uparrow$ & $\Delta p$ $\uparrow$ & Emo-A $\uparrow$ & Emo-S $\uparrow$ &
CLIP-I $\uparrow$ & G. LPIPS $\downarrow$ & G. SSIM $\uparrow$ &
O. LPIPS $\downarrow$ & O. SSIM $\uparrow$ & O. MAE $\downarrow$ \\
\midrule
Qwen-Edit & 0.2996 & 0.2745 & 0.2588 & 0.1843 &
\underline{0.8492} & 0.3698 & 0.5790 & 0.4106 & 0.3049 & 0.1919 \\

\rowcolor{orange!25!yellow!30}\textbf{Qwen-Edit + IECLA (ours)} & \textbf{0.3579} & \textbf{0.3344} & \textbf{0.3081} & \textbf{0.2173} &
\textbf{0.8527} & \textbf{0.3440} & \textbf{0.6137} & \textbf{0.3767} & \textbf{0.3230} & \textbf{0.1733} \\

\quad -w/o attribution & 0.2915 & 0.2735 & 0.2454 & 0.1737 &
0.8394 & 0.3513 & \underline{0.6035} & \underline{0.3851} & \underline{0.3184} & \underline{0.1783} \\

\quad -w/o learning & 0.3046 & 0.2889 & 0.2724 & 0.2017 &
0.8412 & \underline{0.3454} & 0.5877 & 0.4107 & 0.3147 & 0.1918 \\

\quad -w/o agent & \underline{0.3540} & \underline{0.3298} & \underline{0.3024} & \underline{0.2160} &
0.8400 & 0.3543 & 0.5990 & 0.3874 & 0.3171 & 0.1800 \\
\bottomrule
\vspace{-20pt}
\end{tabular}%
}
\endgroup
\end{table}

Qwen-Edit serves as the native baseline. The ablations follow the three modules in
Fig.~\ref{fig:overview}: attribution in (a), contribution learning in (b), and
target-aware execution in (c). IECLA achieves the strongest overall profile,
leading affective metrics and structural measures.
Removing attribution produces the largest affective degradation, confirming that
source-emotion evidence is the primary signal for target-oriented editing. When
the learning module is removed, the attribution output is passed directly from
(a) to (c); the resulting decrease across all affective metrics shows that the
learned contribution representation provides more useful editing evidence than
direct transfer. Removing the agent while retaining attribution yields the
strongest affective scores among the ablations, but still lowers every affective
metric, indicating that the evidence must be converted into a target-aware plan
to realize the full benefit of the framework.

The loss-level ablation further clarifies how the training objective maintains this balance. Focal-sign supervision has the largest affective impact, highlighting the importance of distinguishing supportive from opposing evidence. The ranking terms preserve regional importance and direction, while auxiliary supervision reinforces affective semantics. Although all ablations achieve slightly lower O. LPIPS, they degrade the other preservation measures and affective outcomes. Overall, the complete objective provides the best balance between target-emotion control and structural fidelity.

\vspace{-4pt}
\subsection{Attribution Validation}
\label{sec:reference-validation}
\vspace{-3pt}

\begin{figure}[t]
\centering
\includegraphics[width=\textwidth]{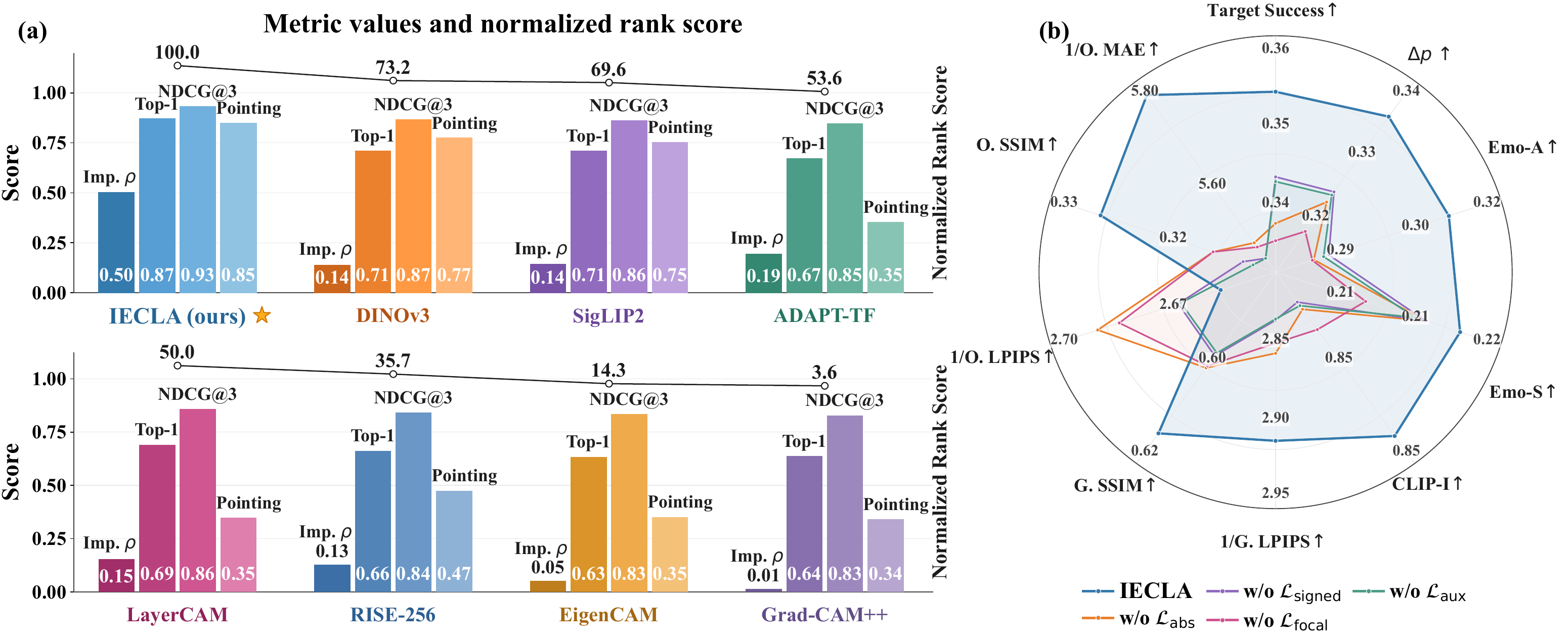}
\vspace{-15pt}
\caption{Attribution validation and loss-function ablation. (a) Metric values and normalized rank scores for representative attribution methods. (b) Ablation of the training objective, where IECLA provides the strongest overall affective and structural profile.}
\vspace{-15pt}
\label{fig:reference-validation}
\end{figure}

Attribution validation assesses whether our predicted regional contributions align with independent external annotations. We evaluate three external VLM references on the same 289 images and 1,032 aggregated SAM regions, with method outputs and aggregation fixed while varying only the reference labels. An independent Human Review provides a complementary perceptual evaluation that is excluded from training and model selection. Figure~\ref{fig:reference-validation}(a) visualizes the Human Review comparison, and Appendix~\ref{app:reference-validation} reports the complete protocol and results.

\FloatBarrier

The same profile appears under all three VLM references: IECLA leads in regional ordering, signed agreement, top-region selection, and localization despite varying competitors. This stability indicates that the result is not an isolated match to one external evaluator. Human Review provides a complementary perceptual test, where IECLA also aligns more closely with both the location and direction of affective evidence. In particular, agreement on direction supports negative contributions as meaningful opposing evidence rather than a by-product of the shared counterfactual teacher.

\vspace{-3pt}
\subsection{User Study}
\label{sec:user-study}
\vspace{-3pt}

A blinded User Study examines whether attribution-guided planning produces perceptually stronger edits without sacrificing visual naturalness. Forty independent reviewers score 120 paired editing tasks across six editors for target expression and naturalness, while 20 additional heatmap tasks assess the perceptual quality of the underlying attribution maps.

Across all six editors, IECLA-guided outputs receive higher mean ratings for both target expression and naturalness. This consistent joint preference indicates that stronger affect does not come at the expense of perceptual naturalness. Together with the comparison in Table~\ref{tab:editing-native}, the common direction across editors supports attribution-guided planning as a transferable editing signal, while differences in gain magnitude reflect how effectively each backbone realizes the localized guidance. The accompanying heatmap evaluation provides complementary perceptual evidence for the readability and focus of the underlying attribution. Overall, the User Study primarily supports the downstream editing utility of IECLA; the complete protocol and detailed results are reported in Appendix~\ref{app:user-study-details}.

\vspace{-3pt}
\section{Conclusion}
\vspace{-3pt}

We presented IECLA, a framework for affective image editing that separates source-emotion attribution from target-specific execution. Its counterfactual contribution predictor localizes signed object-level evidence: positive regions provide an editable focus, while negative regions provide preservation evidence. A target-aware agent then translates this evidence and the desired emotion into localized, content-aware editing operations with explicit preservation constraints. Experiments on the locked attribution benchmark and five external datasets, together with reference validation, show consistent attribution quality across evaluation sources and domains. Across six editing backbones, matched comparisons, ablations, qualitative results, and the User Study further demonstrate improved target-emotion expression with stronger structural preservation. These results establish affective attribution as both an explanation of perceived emotion and an actionable interface for controlling general image editors.

\section*{AI Use Statement}

Generative AI tools supported the development of the conceptual framework, refinement of the experimental methodology, and implementation of experimental code. They were also used to synthesize and analyze related literature, organize references, and improve the clarity and readability of the manuscript. The authors reviewed the resulting text, code, and references and retain full responsibility for all claims, experimental results, and AI-assisted material in the final manuscript.

\section*{Ethics Statement}

This work uses existing image datasets and pretrained models for affective image editing. Human annotation and the user study were conducted with participants' informed consent; their protocols are described in Appendices~\ref{app:human-reference-details} and~\ref{app:user-study-details}. Emotion labels describe perceived affect rather than a depicted individual's internal state. The framework supports creative expression but shares the potential risks of generative image editing, including misleading portrayals and emotional manipulation. Responsible use should respect the privacy of depicted individuals and applicable dataset and model licenses, and disclose image manipulation where authenticity matters.

\section*{Reproducibility Statement}

Section~\ref{sec:method} describes the counterfactual supervision, contribution predictor, and target-aware editing framework, while Section~\ref{sec:experimental-setup} summarizes the experimental setup. Appendix~\ref{app:implementation-details} documents data provenance and disjoint splits, evaluation metrics, reference calibration, training objectives, baseline implementations, and agent and editor configurations. Appendix~\ref{app:agent-template} provides the structured editing-agent template. Appendices~\ref{app:reference-validation} and~\ref{app:user-study-details} describe the external-reference and human-evaluation protocols, while Appendices~\ref{app:loss-ablation} and~\ref{app:stability} provide additional ablations and stability analyses. Code will be released with the final version of the paper. Exact numerical results may vary with stochastic generation and differences in the software and model-serving environment.

\bibliographystyle{iclr2027_conference}
\bibliography{references}

@inproceedings{brooks2023instructpix2pix,
  title={InstructPix2Pix: Learning to Follow Image Editing Instructions},
  author={Brooks, Tim and Holynski, Aleksander and Efros, Alexei A.},
  booktitle={Proceedings of the IEEE/CVF Conference on Computer Vision and Pattern Recognition},
  pages={18392--18403},
  year={2023},
  doi={10.1109/CVPR52729.2023.01764}
}

@article{wu2025qwenimage,
  title={{Qwen-Image} Technical Report},
  author={Wu, Chenfei and Li, Jiahao and Zhou, Jingren and Lin, Junyang and Gao, Kaiyuan and Yan, Kun and Yin, Sheng-ming and Bai, Shuai and Xu, Xiao and Chen, Yilei and others},
  journal={arXiv preprint arXiv:2508.02324},
  year={2025},
  doi={10.48550/arXiv.2508.02324}
}

@misc{blackforestlabs2026flux2,
  title={{FLUX.2 [klein]}: Fast, Efficient Image Generation},
  author={{Black Forest Labs}},
  year={2026},
  howpublished={\url{https://bfl.ai/models/flux-2-klein}}
}

@article{longcat2025image,
  title={{LongCat-Image} Technical Report},
  author={{Meituan LongCat Team} and Ma, Hanghang and Tan, Haoxian and Huang, Jiale and Wu, Junqiang and He, Jun-Yan and Gao, Lishuai and Xiao, Songlin and Wei, Xiaoming and Ma, Xiaoqi and Cai, Xunliang and Guan, Yayong and Hu, Jie},
  journal={arXiv preprint arXiv:2512.07584},
  year={2025},
  doi={10.48550/arXiv.2512.07584}
}

@article{wu2025omnigen2,
  title={{OmniGen2}: Towards Instruction-Aligned Multimodal Generation},
  author={Wu, Chenyuan and Zheng, Pengfei and Yan, Ruiran and Xiao, Shitao and Luo, Xin and Wang, Yueze and Li, Wanli and Jiang, Xiyan and Liu, Yexin and Zhou, Junjie and others},
  journal={arXiv preprint arXiv:2506.18871},
  year={2025},
  doi={10.48550/arXiv.2506.18871}
}

@article{cai2025hidreami1,
  title={{HiDream-I1}: A High-Efficient Image Generative Foundation Model with Sparse Diffusion Transformer},
  author={Cai, Qi and Chen, Jingwen and Chen, Yang and Li, Yehao and Long, Fuchen and Pan, Yingwei and Qiu, Zhaofan and Zhang, Yiheng and Gao, Fengbin and Xu, Peihan and others},
  journal={arXiv preprint arXiv:2505.22705},
  year={2025},
  doi={10.48550/arXiv.2505.22705}
}

@inproceedings{selvaraju2017gradcam,
  title={Grad-CAM: Visual Explanations from Deep Networks via Gradient-Based Localization},
  author={Selvaraju, Ramprasaath R. and Cogswell, Michael and Das, Abhishek and Vedantam, Ramakrishna and Parikh, Devi and Batra, Dhruv},
  booktitle={Proceedings of the IEEE International Conference on Computer Vision},
  pages={618--626},
  year={2017},
  doi={10.1109/ICCV.2017.74}
}

@inproceedings{liu2024groundingdino,
  title={Grounding DINO: Marrying DINO with Grounded Pre-Training for Open-Set Object Detection},
  author={Liu, Shilong and Zeng, Zhaoyang and Ren, Tianhe and Li, Feng and Zhang, Hao and Yang, Jie and Jiang, Qing and Li, Chunyuan and Yang, Jianwei and Su, Hang and Zhu, Jun and Zhang, Lei},
  booktitle={European Conference on Computer Vision},
  pages={38--55},
  year={2024},
  doi={10.1007/978-3-031-72970-6_3}
}

@inproceedings{kirillov2023sam,
  title={Segment Anything},
  author={Kirillov, Alexander and Mintun, Eric and Ravi, Nikhila and Mao, Hanzi and Rolland, Chloe and Gustafson, Laura and Xiao, Tete and Whitehead, Spencer and Berg, Alexander C. and Lo, Wan-Yen and Doll{\'a}r, Piotr and Girshick, Ross},
  booktitle={Proceedings of the IEEE/CVF International Conference on Computer Vision},
  pages={4015--4026},
  year={2023},
  doi={10.1109/ICCV51070.2023.00371}
}

@inproceedings{achlioptas2021artemis,
  title={ArtEmis: Affective Language for Visual Art},
  author={Achlioptas, Panos and Ovsjanikov, Maks and Haydarov, Kilichbek and Elhoseiny, Mohamed and Guibas, Leonidas},
  booktitle={Proceedings of the IEEE/CVF Conference on Computer Vision and Pattern Recognition},
  pages={11564--11574},
  year={2021},
  doi={10.1109/CVPR46437.2021.01140}
}

@inproceedings{yang2023emoset,
  title={EmoSet: A Large-Scale Visual Emotion Dataset with Rich Attributes},
  author={Yang, Jingyuan and Huang, Qirui and Ding, Tingting and Lischinski, Dani and Cohen-Or, Daniel and Huang, Hui},
  booktitle={Proceedings of the IEEE/CVF International Conference on Computer Vision},
  pages={20326--20337},
  year={2023},
  doi={10.1109/ICCV51070.2023.01864}
}

@inproceedings{lin2014coco,
  title={Microsoft {COCO}: Common Objects in Context},
  author={Lin, Tsung-Yi and Maire, Michael and Belongie, Serge and Hays, James and Perona, Pietro and Ramanan, Deva and Doll{\'a}r, Piotr and Zitnick, C. Lawrence},
  booktitle={European Conference on Computer Vision},
  pages={740--755},
  year={2014},
  doi={10.1007/978-3-319-10602-1_48}
}

@inproceedings{you2016building,
  title={Building a Large Scale Dataset for Image Emotion Recognition: The Fine Print and The Benchmark},
  author={You, Quanzeng and Luo, Jiebo and Jin, Hailin and Yang, Jianchao},
  booktitle={Proceedings of the AAAI Conference on Artificial Intelligence},
  year={2016},
  doi={10.1609/aaai.v30i1.9987}
}

@inproceedings{zhang2025emoart,
  title={EmoArt: A Multidimensional Dataset for Emotion-Aware Artistic Generation},
  author={Zhang, Cheng and Xie, Hongxia and Wen, Bin and Zuo, Songhan and Zhang, Ruoxuan and Cheng, Wen-Huang},
  booktitle={Proceedings of the 33rd ACM International Conference on Multimedia},
  pages={12644--12650},
  year={2025},
  doi={10.1145/3746027.3758201}
}

@inproceedings{zhang2023controlnet,
  title={Adding Conditional Control to Text-to-Image Diffusion Models},
  author={Zhang, Lvmin and Rao, Anyi and Agrawala, Maneesh},
  booktitle={Proceedings of the IEEE/CVF International Conference on Computer Vision},
  pages={3813--3824},
  year={2023},
  doi={10.1109/ICCV51070.2023.00355}
}

@inproceedings{ribeiro2016lime,
  title={Why Should I Trust You?: Explaining the Predictions of Any Classifier},
  author={Ribeiro, Marco Tulio and Singh, Sameer and Guestrin, Carlos},
  booktitle={Proceedings of the 2016 Conference of the North American Chapter of the Association for Computational Linguistics: Demonstrations},
  pages={97--101},
  year={2016},
  doi={10.18653/v1/N16-3020}
}

@inproceedings{fong2017meaningful,
  title={Interpretable Explanations of Black Boxes by Meaningful Perturbation},
  author={Fong, Ruth C. and Vedaldi, Andrea},
  booktitle={Proceedings of the IEEE International Conference on Computer Vision},
  pages={3449--3457},
  year={2017},
  doi={10.1109/ICCV.2017.371}
}

@inproceedings{peng2015mixedbag,
  title={A Mixed Bag of Emotions: Model, Predict, and Transfer Emotion Distributions},
  author={Peng, Kuan-Chuan and Chen, Tsuhan and Sadovnik, Amir and Gallagher, Andrew},
  booktitle={2015 IEEE Conference on Computer Vision and Pattern Recognition (CVPR)},
  pages={860--868},
  year={2015},
  doi={10.1109/CVPR.2015.7298687}
}

@article{kosti2020emotic,
  title={Context Based Emotion Recognition using {EMOTIC} Dataset},
  author={Kosti, Ronak and Alvarez, Jose M. and Recasens, Adria and Lapedriza, Agata},
  journal={IEEE Transactions on Pattern Analysis and Machine Intelligence},
  volume={42},
  number={11},
  pages={2755--2766},
  year={2020},
  doi={10.1109/TPAMI.2019.2916866}
}

@inproceedings{sundararajan2017integrated,
  title={Axiomatic Attribution for Deep Networks},
  author={Sundararajan, Mukund and Taly, Ankur and Yan, Qiqi},
  booktitle={Proceedings of the 34th International Conference on Machine Learning},
  series={Proceedings of Machine Learning Research},
  volume={70},
  pages={3319--3328},
  year={2017},
  publisher={PMLR}
}

@inproceedings{petsiuk2018rise,
  title={{RISE}: Randomized Input Sampling for Explanation of Black-box Models},
  author={Petsiuk, Vitali and Das, Abir and Saenko, Kate},
  booktitle={British Machine Vision Conference 2018},
  pages={151},
  year={2018},
  publisher={BMVA Press}
}

@inproceedings{goyal2019counterfactual,
  title={Counterfactual Visual Explanations},
  author={Goyal, Yash and Wu, Ziyan and Ernst, Jan and Batra, Dhruv and Parikh, Devi and Lee, Stefan},
  booktitle={Proceedings of the 36th International Conference on Machine Learning},
  series={Proceedings of Machine Learning Research},
  volume={97},
  pages={2376--2384},
  year={2019},
  publisher={PMLR}
}

@inproceedings{meng2022sdedit,
  title={{SDEdit}: Guided Image Synthesis and Editing with Stochastic Differential Equations},
  author={Meng, Chenlin and He, Yutong and Song, Yang and Song, Jiaming and Wu, Jiajun and Zhu, Jun-Yan and Ermon, Stefano},
  booktitle={International Conference on Learning Representations},
  year={2022},
  url={https://arxiv.org/abs/2108.01073}
}

@article{geng2023instructdiffusion,
  title={InstructDiffusion: A Generalist Modeling Interface for Vision Tasks},
  author={Geng, Zigang and Yang, Binxin and Hang, Tiankai and Li, Chen and Gu, Shuyang and Zhang, Ting and Bao, Jianmin and Zhang, Zheng and Hu, Han and Chen, Dong and Guo, Baining},
  journal={arXiv preprint arXiv:2309.03895},
  year={2023},
  doi={10.48550/arXiv.2309.03895}
}

@misc{li2023blipdiffusionpretrainedsubjectrepresentation,
      title={BLIP-Diffusion: Pre-trained Subject Representation for Controllable Text-to-Image Generation and Editing}, 
      author={Dongxu Li and Junnan Li and Steven C. H. Hoi},
      year={2023},
      eprint={2305.14720},
      archivePrefix={arXiv},
      primaryClass={cs.CV},
      url={https://arxiv.org/abs/2305.14720}, 
}

@inproceedings{fu2022language,
  title={Language-Driven Artistic Style Transfer},
  author={Fu, Tsu-Jui and Wang, Xin Eric and Wang, William Yang},
  booktitle={European Conference on Computer Vision},
  pages={717--734},
  year={2022}
}

@inproceedings{weng2023aif,
  title={Affective Image Filter: Reflecting Emotions from Text to Images},
  author={Weng, Shuchen and Zhang, Peixuan and Chang, Zheng and Wang, Xinlong and Li, Si and Shi, Boxin},
  booktitle={Proceedings of the IEEE/CVF International Conference on Computer Vision},
  pages={10810--10819},
  year={2023}
}

@inproceedings{hertz2023prompt,
  title={Prompt-to-Prompt Image Editing with Cross Attention Control},
  author={Hertz, Amir and Mokady, Ron and Tenenbaum, Jay and Aberman, Kfir and Pritch, Yael and Cohen-Or, Daniel},
  booktitle={International Conference on Learning Representations},
  year={2023},
  url={https://arxiv.org/abs/2208.01626}
}

@inproceedings{avrahami2022blended,
  title={Blended Diffusion for Text-driven Editing of Natural Images},
  author={Avrahami, Omri and Lischinski, Dani and Fried, Ohad},
  booktitle={2022 IEEE/CVF Conference on Computer Vision and Pattern Recognition (CVPR)},
  pages={18187--18197},
  year={2022},
  doi={10.1109/CVPR52688.2022.01767}
}

@inproceedings{zhang2023magicbrush,
  title={MagicBrush: A Manually Annotated Dataset for Instruction-Guided Image Editing},
  author={Zhang, Kai and Mo, Lingbo and Chen, Wenhu and Sun, Huan and Su, Yu},
  booktitle={Advances in Neural Information Processing Systems 36},
  pages={31428--31449},
  year={2023},
  doi={10.52202/075280-1365}
}

@article{tschannen2025siglip2,
  title={SigLIP 2: Multilingual Vision-Language Encoders with Improved Semantic Understanding, Localization, and Dense Features},
  author={Tschannen, Michael and Gritsenko, Alexey and Wang, Xiao and Naeem, Muhammad Ferjad and Alabdulmohsin, Ibrahim and Parthasarathy, Nikhil and Evans, Talfan and Beyer, Lucas and Xia, Ye and Mustafa, Basil and H{\'e}naff, Olivier and Harmsen, Jeremiah and Steiner, Andreas and Zhai, Xiaohua},
  journal={arXiv preprint arXiv:2502.14786},
  year={2025},
  doi={10.48550/arXiv.2502.14786}
}

@article{adebayo2018sanity,
  title={Sanity Checks for Saliency Maps},
  author={Adebayo, Julius and Gilmer, Justin and Muelly, Michael and Goodfellow, Ian and Hardt, Moritz and Kim, Been},
  journal={arXiv preprint arXiv:1810.03292},
  year={2018},
  doi={10.48550/arXiv.1810.03292}
}

@article{jiang2021layercam,
  title={LayerCAM: Exploring Hierarchical Class Activation Maps for Localization},
  author={Jiang, Peng-Tao and Zhang, Chang-Bin and Hou, Qibin and Cheng, Ming-Ming and Wei, Yunchao},
  journal={IEEE Transactions on Image Processing},
  volume={30},
  pages={5875--5888},
  year={2021},
  doi={10.1109/TIP.2021.3089943}
}

@article{doulfoukar2024emocam,
  title={EmoCAM: Toward Understanding What Drives {CNN}-based Emotion Recognition},
  author={Doulfoukar, Youssef and Mertens, Laurent and Vennekens, Joost},
  journal={arXiv preprint arXiv:2407.14314},
  year={2024},
  doi={10.48550/arXiv.2407.14314}
}

@inproceedings{yang2025emoedit,
  title={Emoedit: Evoking emotions through image manipulation},
  author={Yang, Jingyuan and Feng, Jiawei and Luo, Weibin and Lischinski, Dani and Cohen-Or, Daniel and Huang, Hui},
  booktitle={2025 IEEE/CVF Conference on Computer Vision and Pattern Recognition (CVPR)},
  pages={24690--24699},
  year={2025},
  organization={IEEE}
}

@article{ye2025moodifier,
  title={Moodifier: {MLLM}-Enhanced Emotion-Driven Image Editing},
  author={Ye, Jiarong and Huang, Sharon X.},
  journal={arXiv preprint arXiv:2507.14024},
  year={2025},
  doi={10.48550/arXiv.2507.14024}
}

@article{mao2025emoagent,
  title={EmoAgent: A Multi-Agent Framework for Diverse Affective Image Manipulation},
  author={Mao, Qi and Hu, Haobo and He, Yujie and Gao, Difei and Chen, Haokun and Jin, Libiao},
  journal={arXiv preprint arXiv:2503.11290},
  year={2025},
  doi={10.48550/arXiv.2503.11290}
}

@article{man2025vaeer,
  title={{VAEER}: Visual Attention-Inspired Emotion Elicitation Reasoning},
  author={Man, Fanhang and Chen, Xiaoyue and Wang, Huandong and Zhao, Baining and Li, Han and Chen, Xinlei},
  journal={arXiv preprint arXiv:2505.24342},
  year={2025},
  doi={10.48550/arXiv.2505.24342}
}

@article{zhang2026emokgedit,
  title={EmoKGEdit: Training-free Affective Injection via Visual Cue Transformation},
  author={Zhang, Jing and Fan, Bingjie},
  journal={arXiv preprint arXiv:2601.12326},
  year={2026},
  doi={10.48550/arXiv.2601.12326}
}

@article{yin2026mood,
  title={MooD: Perception-Enhanced Efficient Affective Image Editing via Continuous Valence--Arousal Modeling},
  author={Yin, Xinyi and Wang, Yiduo and Hu, Tingqi and Si, Meicong and Shi, Yunyun and Chen, Shi and Wang, Hao and Xue, Junxiao and Wu, Xuecheng},
  journal={arXiv preprint arXiv:2605.02521},
  year={2026},
  doi={10.48550/arXiv.2605.02521}
}

@article{simeoni2025dinov3,
  title={{DINOv3}},
  author={Sim{\'e}oni, Oriane and Vo, Huy V. and Seitzer, Maximilian and Baldassarre, Federico and Oquab, Maxime and Jose, Cijo and Khalidov, Vasil and Szafraniec, Marc and Yi, Seungeun and Ramamonjisoa, Micha{\"e}l and Massa, Francisco and Haziza, Daniel and Wehrstedt, Luca and Wang, Jianyuan and Darcet, Timoth{\'e}e and Moutakanni, Th{\'e}o and Sentana, Leonel and Roberts, Claire and Vedaldi, Andrea and Tolan, Jamie and Brandt, John and Couprie, Camille and Mairal, Julien and J{\'e}gou, Herv{\'e} and Labatut, Patrick and Bojanowski, Piotr},
  journal={arXiv preprint arXiv:2508.10104},
  year={2025},
  doi={10.48550/arXiv.2508.10104}
}

@inproceedings{desai2020ablationcam,
  title={Ablation-CAM: Visual Explanations for Deep Convolutional Network via Gradient-free Localization},
  author={Desai, Saurabh and Ramaswamy, Harish G.},
  booktitle={2020 IEEE Winter Conference on Applications of Computer Vision},
  pages={972--980},
  year={2020},
  doi={10.1109/WACV45572.2020.9093360}
}

@inproceedings{chattopadhay2018gradcampp,
  title={Grad-CAM++: Generalized Gradient-Based Visual Explanations for Deep Convolutional Networks},
  author={Chattopadhay, Aditya and Sarkar, Anirban and Howlader, Prantik and Balasubramanian, Vineeth N.},
  booktitle={2018 IEEE Winter Conference on Applications of Computer Vision},
  pages={839--847},
  year={2018},
  doi={10.1109/WACV.2018.00097}
}

@inproceedings{muhammad2020eigencam,
  title={Eigen-CAM: Class Activation Map Using Principal Components},
  author={Muhammad, Mohammed Bany and Yeasin, Mohammed},
  booktitle={2020 International Joint Conference on Neural Networks},
  pages={1--7},
  year={2020},
  doi={10.1109/IJCNN48605.2020.9206626}
}

@article{draelos2020use,
  title={Use HiResCAM instead of Grad-CAM for faithful explanations of convolutional neural networks},
  author={Draelos, Rachel Lea and Carin, Lawrence},
  journal={arXiv preprint arXiv:2011.08891},
  year={2020}
}

@inproceedings{radford2021clip,
  title={Learning Transferable Visual Models from Natural Language Supervision},
  author={Radford, Alec and Kim, Jong Wook and Hallacy, Chris and Ramesh, Aditya and Goh, Gabriel and Agarwal, Sandhini and Sastry, Girish and Askell, Amanda and Mishkin, Pamela and Clark, Jack and Krueger, Gretchen and Sutskever, Ilya},
  booktitle={Proceedings of the 38th International Conference on Machine Learning},
  series={Proceedings of Machine Learning Research},
  volume={139},
  pages={8748--8763},
  year={2021},
  publisher={PMLR}
}

@article{yao2026adapt,
  title={{ADAPT}: Attention Dynamics Alignment with Preference Tuning for Faithful {MLLMs}},
  author={Yao, Zhiyuan and Fu, Zheren and Zheng, Zhixiao and Li, Jiajun and Tu, Yi and Mao, Zhendong},
  journal={arXiv preprint arXiv:2606.31054},
  year={2026},
  doi={10.48550/arXiv.2606.31054}
}

@article{guo2025emoverse,
  title={Emoverse: A mllms-driven emotion representation dataset for interpretable visual emotion analysis},
  author={Guo, Yijie and Hong, Dexiang and Chen, Weidong and She, Zihan and Ye, Cheng and Chang, Xiaojun and Mao, Zhendong},
  journal={arXiv preprint arXiv:2511.12554},
  year={2025}
}

@inproceedings{fu2024guiding,
  title={Guiding instruction-based image editing via multimodal large language models},
  author={Fu, Tsu-Jui and Hu, Wenze and Du, Xianzhi and Wang, William and Yang, Yinfei and Gan, Zhe},
  booktitle={International Conference on Learning Representations},
  volume={2024},
  pages={54820--54833},
  year={2024}
}

@inproceedings{yang2024cross,
  title={Cross-modal contextualized diffusion models for text-guided visual generation and editing},
  author={Yang, Ling and Zhang, Zhilong and Yu, Zhaochen and Liu, Jingwei and Xu, Minkai and Ermon, Stefano and Cui, Bin},
  booktitle={International Conference on Learning Representations},
  volume={2024},
  pages={13972--13997},
  year={2024}
}

@inproceedings{wei2025omniedit,
  title={Omniedit: Building image editing generalist models through specialist supervision},
  author={Wei, Cong and Xiong, Zheyang and Ren, Weiming and Du, Xeron and Zhang, Ge and Chen, Wenhu},
  booktitle={International Conference on Learning Representations},
  volume={2025},
  pages={259--271},
  year={2025}
}

@inproceedings{yuan2026coemogen,
  title={Coemogen: Towards semantically-coherent and scalable emotional image content generation},
  author={Yuan, Kaishen and Zhang, Yuting and Gao, Shang and Zhu, Yijie and Chen, Wenshuo and Yue, Yutao},
  booktitle={International Conference on Learning Representations},
  volume={2026},
  pages={6341--6361},
  year={2026}
}

@inproceedings{ye2024dual,
  title={Dual-path collaborative generation network for emotional video captioning},
  author={Ye, Cheng and Chen, Weidong and Li, Jingyu and Zhang, Lei and Mao, Zhendong},
  booktitle={Proceedings of the 32nd ACM International Conference on Multimedia},
  pages={496--505},
  year={2024}
}

@article{hong2026emostyle,
  title={EmoStyle: Affective Conditioning of Style-Specialist Experts for Emotional Image Generation},
  author={Hong, Dexiang and Guo, Yijie and Chen, Weidong and Liu, Xinyan and Zou, Zixuan and Mao, Zhendong and Zhang, Yongdong},
  journal={arXiv preprint arXiv:2607.10165},
  year={2026}
}

@article{wang2026multi,
  title={A multi-agent framework with structured reasoning and reflective refinement for multimodal empathetic response generation},
  author={Wang, Liping and Ye, Cheng and Chen, Weidong and Song, Peipei and Hu, Bo and Mao, Zhendong},
  journal={arXiv preprint arXiv:2604.18988},
  year={2026}
}

@article{chen2026face,
  title={Face-net: Factual calibration and emotion augmentation for retrieval-enhanced emotional video captioning},
  author={Chen, Weidong and Ye, Cheng and Mao, Zhendong and Song, Peipei and Liu, Xinyan and Zhang, Lei and Chang, Xiaojun and Zhang, Yongdong},
  journal={arXiv preprint arXiv:2603.17455},
  year={2026}
}

@inproceedings{zhang2026creatidesign,
  title={Creatidesign: A unified multi-conditional diffusion transformer for creative graphic design},
  author={Zhang, Hui and Hong, Dexiang and Yang, Maoke and Cheng, Yutao and Zhang, Zhao and Chen, Weidong and Shao, Jie and Wu, Xinglong and Wu, Zuxuan and Jiang, Yu-Gang},
  booktitle={International Conference on Learning Representations},
  volume={2026},
  pages={111201--111214},
  year={2026}
}

@article{zhang2025creatiposter,
  title={Creatiposter: Towards editable and controllable multi-layer graphic design generation},
  author={Zhang, Zhao and Cheng, Yutao and Hong, Dexiang and Yang, Maoke and Shi, Gonglei and Ma, Lei and Zhang, Hui and Shao, Jie and Wu, Xinglong},
  journal={arXiv preprint arXiv:2506.10890},
  year={2025}
}

\clearpage
\appendix
\raggedbottom
\newcommand{\appendixtablesetup}{%
  \centering
  \scriptsize
  \setlength{\tabcolsep}{2.3pt}%
  \renewcommand{\arraystretch}{0.94}%
  \setlength{\abovecaptionskip}{2pt}%
  \setlength{\belowcaptionskip}{3pt}%
}
\newcommand{\appendixcontentsentry}[2]{%
  \noindent\hyperref[#1]{\textbf{\ref*{#1}\quad #2}}\dotfill\textbf{\pageref*{#1}}\par
  \vspace{4pt}%
}
\newcommand{\appendixsubcontentsentry}[2]{%
  \noindent\hspace{1.8em}\hyperref[#1]{\ref*{#1}\quad #2}\dotfill\pageref*{#1}\par
  \vspace{2pt}%
}

\begin{center}
{\LARGE\bfseries Supplementary Material}
\end{center}
\vspace{1em}

\section*{Contents}
\appendixcontentsentry{app:additional-editing-results}{Qualitative Results}
\appendixcontentsentry{app:editing-baseline-comparison}{Editing Baselines}
\appendixcontentsentry{app:cross-dataset-details}{Cross-Dataset Results}
\appendixcontentsentry{app:reference-validation}{Reference Validation}
\appendixsubcontentsentry{app:vlm-reference-details}{VLM Reference}
\appendixsubcontentsentry{app:human-reference-details}{Human Reference}
\appendixcontentsentry{app:user-study-details}{User Study}
\appendixcontentsentry{app:loss-ablation}{Additional Ablations}
\appendixcontentsentry{app:stability}{Stability Analysis}
\appendixcontentsentry{app:frozen-teacher}{Teacher Replay and Predictor Comparison}
\appendixcontentsentry{app:agent-template}{Agent Template}
\appendixcontentsentry{app:implementation-details}{Implementation Details}

\clearpage
\section{Qualitative Results}
\label{app:additional-editing-results}

\begin{figure}[H]
\centering
\includegraphics[width=\textwidth, pagebox=cropbox]{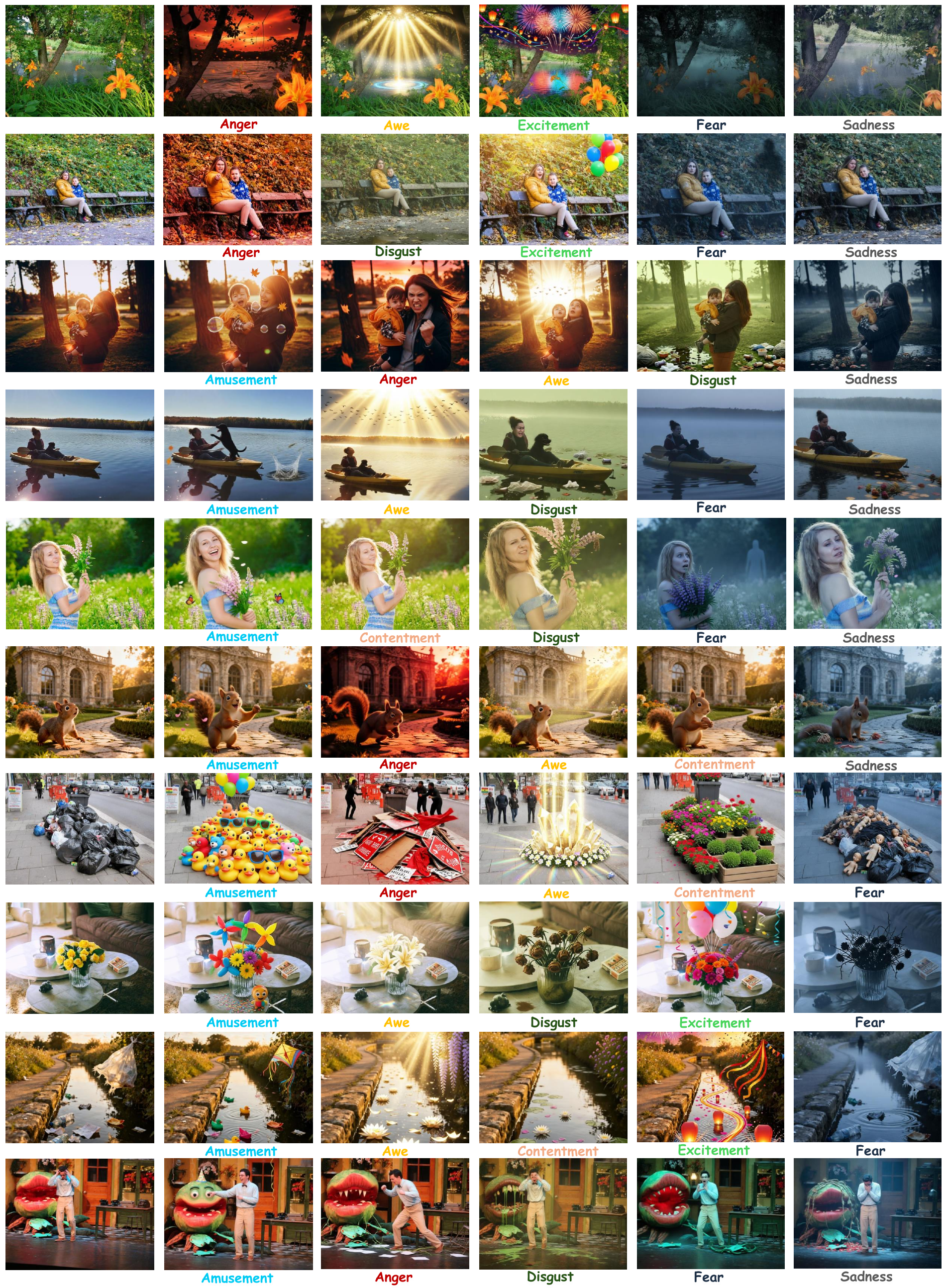}
\vspace{-8pt}
\caption{Additional qualitative results with IECLA-guided Qwen-Edit. Each row presents a source image followed by edits toward different target emotions, showing diverse affective changes with consistent source content and composition.}
\label{fig:additional-editing-results}
\end{figure}

Figure~\ref{fig:additional-editing-results} complements the examples in Figure~\ref{fig:teaser} with additional IECLA-guided Qwen-Edit outputs. The examples span people, animals, objects, and scene-centric images. Across these settings, the attributed guidance produces target-specific changes through localized semantic cues and scene appearance while retaining recognizable source content and composition.

\begin{figure}[H]
\centering
\includegraphics[width=\textwidth,height=0.78\textheight,keepaspectratio,pagebox=cropbox]{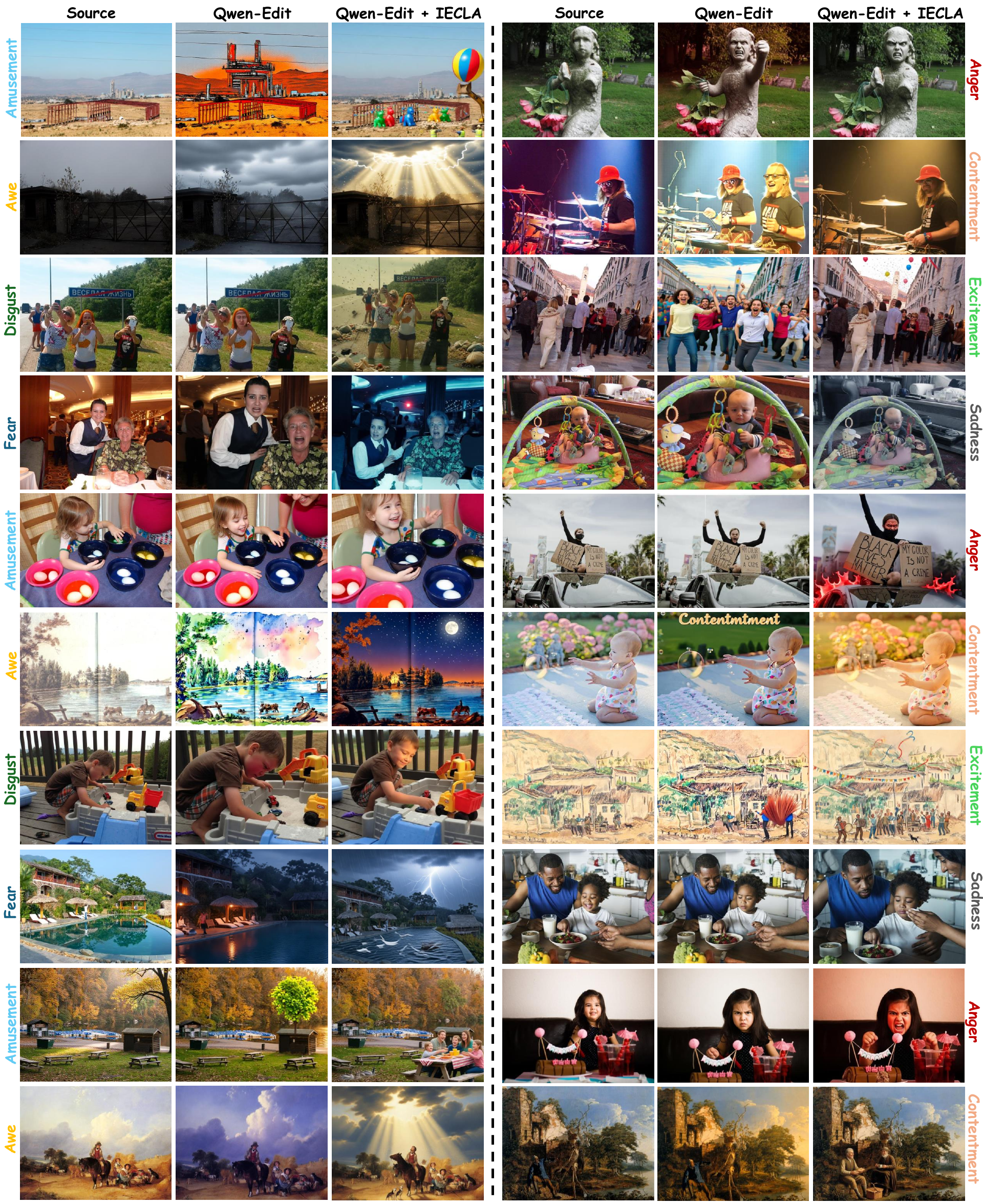}
\vspace{-8pt}
\caption{Additional comparisons between native Qwen-Edit and IECLA-guided editing. IECLA strengthens target-emotion expression while preserving scene structure across diverse images.}
\label{fig:additional-native-comparison}
\end{figure}

Figure~\ref{fig:additional-native-comparison} further compares native Qwen-Edit with the same editor guided by IECLA. Across varied source content and target emotions, attribution-guided editing introduces clearer emotion-specific cues while better preserving the original subject and scene organization.

Taken together, the examples expose two recurring failure modes of native editing. When a source image already carries cues close to contentment, the native editor may introduce text or decorative objects even when a small chromatic or illumination adjustment would suffice. IECLA instead uses positive signed counterfactual contributions as editable evidence and negative contributions as preservation evidence, allowing the agent to make a restrained change while retaining the original composition. Several contentment examples follow this pattern: the guided outputs adjust color and light without adding unrelated content.

For targets such as fear, the native editor often darkens the background or adds red illumination as isolated operations. These cues do not necessarily form a coherent relation with the foreground and can leave the perceived emotion weak or ambiguous. IECLA can route evidence across the subject and its surrounding context, so expression, lighting, and background cues are changed jointly when supported by the image. The same pattern appears in face-dominant scenes, where editing only the expression misses contextual evidence, and in artistic images, where IECLA changes palette or atmosphere while preserving composition. These cases illustrate how signed attribution supplies relational editing evidence rather than a generic instruction to modify the most salient region.

\FloatBarrier
\section{Editing Baselines}
\label{app:editing-baseline-comparison}

This appendix compares IECLA with representative affective and general-purpose image editing methods on the same test set used for the main editing tables. The comparison includes global diffusion editing (SDEdit~\citep{meng2022sdedit}), instruction-based editing (InsDiff, i.e., InstructDiffusion~\citep{geng2023instructdiffusion}), spatial control (ControlNet~\citep{zhang2023controlnet}), subject-preserving editing (BLIP-Diffusion~\citep{li2023blipdiffusionpretrainedsubjectrepresentation}), and affective style transfer (CLVA~\citep{fu2022language} and AIF~\citep{weng2023aif}), together with EmoEdit~\citep{yang2025emoedit}. All methods are evaluated on the 680 source images and 5,440 source--target tasks used in the main editing comparison; the Qwen-Edit + IECLA row uses the same IECLA-guided Qwen-Edit configuration as the main tables. Since absolute scores depend on the editing backbone, this single-backbone comparison is reported here, while Table~\ref{tab:editing-native} presents matched native-versus-guided results across six editors.

Table~\ref{tab:editing-baseline-comparison} shows the tension between emotion control and structural preservation across prior approaches. Qwen-Edit + IECLA leads in five of six evaluations, indicating stronger target-emotion expression alongside close agreement with the source image. EmoEdit ranks second on LPIPS, while AIF leads in CLIP-I; the latter metric is also high for CLVA despite its low emotion scores. Thus, high image similarity alone does not imply effective emotion control, whereas IECLA performs well on both affective and preservation measures.

\begin{table}[H]
\appendixtablesetup
\caption{Comparison with affective and general-purpose editing baselines on the main editing test set. IECLA provides the strongest joint emotion improvement and structure retention.}
\label{tab:editing-baseline-comparison}
\begin{tabular}{lrrrrrr}
\toprule
Method & PSNR$\uparrow$ & SSIM$\uparrow$ & LPIPS$\downarrow$ & CLIP-I$\uparrow$ & Emo-A$\uparrow$ & Emo-S$\uparrow$ \\
\midrule
SDEdit~\citep{meng2022sdedit} & 14.1547 & 0.3820 & 0.3841 & 0.7481 & 0.2050 & 0.0929 \\
InsDiff~\citep{geng2023instructdiffusion} & 9.8607 & 0.3870 & 0.3999 & 0.7388 & 0.0551 & 0.0537 \\
ControlNet~\citep{zhang2023controlnet} & 12.2646 & 0.2370 & 0.4893 & 0.6768 & 0.1860 & 0.1659 \\
BLIP-Diffusion~\citep{li2023blipdiffusionpretrainedsubjectrepresentation} & 8.7242 & 0.3260 & 0.5514 & 0.8169 & 0.0676 & 0.0190 \\
CLVA~\citep{fu2022language} & 12.1975 & 0.4960 & 0.3963 & \underline{0.8572} & 0.0379 & 0.0138 \\
AIF~\citep{weng2023aif} & 16.1320 & 0.6080 & 0.4147 & \textbf{0.9932} & 0.0812 & 0.0345 \\
EmoEdit~\citep{yang2025emoedit} & \underline{18.1564} & \underline{0.6090} & \underline{0.3459} & 0.8413 & \underline{0.2647} & \underline{0.2149} \\
\midrule
\rowcolor{orange!25!yellow!30}\textbf{Qwen-Edit + IECLA (ours)} & \textbf{20.3046} & \textbf{0.6137} & \textbf{0.3440} & 0.8527 & \textbf{0.3081} & \textbf{0.2173} \\
\bottomrule
\end{tabular}
\end{table}

\FloatBarrier
\section{Cross-Dataset Results}
\label{app:cross-dataset-details}

The five external datasets are examined separately under the attribution protocol summarized in Table~\ref{tab:cross-dataset-attribution-summary}. All tables use the same frozen counterfactual teacher and Grounding DINO and SAM candidate regions. Ranking and localization columns use image-macro aggregation, whereas Sign Macro-F1 and regression columns pool all valid candidate regions; images without valid regions are removed independently within each dataset. Bold and underlined entries mark the best and second-best results, while ``--'' denotes metrics undefined for nonnegative saliency outputs.

\begin{table}[H]
\appendixtablesetup
\caption{Cross-dataset attribution results on COCO (612 images, 5,008 masks).}
\label{tab:cross-coco}
\begin{tabular}{lrrcccccccc}
\toprule
Method & Images & Masks & Imp. $\rho\uparrow$ & Signed $\rho\uparrow$ & Top-1$\uparrow$ & NDCG@3$\uparrow$ & Macro-F1$\uparrow$ & MAE$\downarrow$ & RMSE$\downarrow$ & $R^2\uparrow$ \\
\midrule
\rowcolor{gray!15}\multicolumn{11}{l}{\textit{Heatmap attribution methods}} \\
Ablation-CAM & 612 & 5008 & -0.0632 & -- & 0.1765 & 0.5035 & -- & -- & -- & -- \\
Grad-CAM & 612 & 5008 & 0.0740 & -- & 0.2320 & 0.5814 & -- & -- & -- & -- \\
Direct attention & 612 & 5008 & 0.0507 & -- & 0.2141 & 0.5541 & -- & -- & -- & -- \\
Grad-CAM++ & 612 & 5008 & 0.0697 & -- & 0.2190 & 0.5737 & -- & -- & -- & -- \\
RISE-256 & 612 & 5008 & 0.1688 & -- & 0.2467 & 0.5986 & -- & -- & -- & -- \\
ADAPT-TF & 612 & 5008 & -0.0061 & -- & 0.1912 & 0.5125 & -- & -- & -- & -- \\
EigenCAM & 612 & 5008 & 0.1102 & -- & 0.2533 & 0.5950 & -- & -- & -- & -- \\
HiResCAM & 612 & 5008 & 0.1840 & -- & 0.2565 & 0.6413 & -- & -- & -- & -- \\
LayerCAM & 612 & 5008 & 0.2162 & -- & 0.2843 & 0.6554 & -- & -- & -- & -- \\
\midrule
\rowcolor{gray!15}\multicolumn{11}{l}{\textit{Vision-encoder contribution methods}} \\
CLIP & 612 & 5008 & 0.3549 & 0.1995 & 0.4444 & 0.7530 & 0.4669 & 0.1617 & 0.2458 & 0.0102 \\
DINOV3 & 612 & 5008 & 0.3853 & 0.1977 & 0.4248 & 0.7588 & 0.4950 & 0.1459 & 0.2394 & 0.0610 \\
SigLIP2 & 612 & 5008 & \underline{0.4038} & \underline{0.3008} & \underline{0.4641} & \underline{0.7800} & \underline{0.5442} & \underline{0.1306} & \underline{0.2078} & \underline{0.2924} \\
\rowcolor{orange!25!yellow!30}\textbf{IECLA (ours)} & 612 & 5008 & \textbf{0.5833} & \textbf{0.5558} & \textbf{0.5735} & \textbf{0.8656} & \textbf{0.7674} & \textbf{0.0757} & \textbf{0.1545} & \textbf{0.6087} \\
\bottomrule
\end{tabular}%
\end{table}

\begin{table}[H]
\appendixtablesetup
\caption{Cross-dataset attribution results on EmoSet (915 images, 6,304 masks).}
\label{tab:cross-emoset}
\begin{tabular}{lrrcccccccc}
\toprule
Method & Images & Masks & Imp. $\rho\uparrow$ & Signed $\rho\uparrow$ & Top-1$\uparrow$ & NDCG@3$\uparrow$ & Macro-F1$\uparrow$ & MAE$\downarrow$ & RMSE$\downarrow$ & $R^2\uparrow$ \\
\midrule
\rowcolor{gray!15}\multicolumn{11}{l}{\textit{Heatmap attribution methods}} \\
Ablation-CAM & 915 & 6304 & -0.0195 & -- & 0.2634 & 0.5877 & -- & -- & -- & -- \\
Grad-CAM & 915 & 6304 & 0.0089 & -- & 0.2710 & 0.6188 & -- & -- & -- & -- \\
Direct attention & 915 & 6304 & 0.0395 & -- & 0.2667 & 0.6307 & -- & -- & -- & -- \\
Grad-CAM++ & 915 & 6304 & 0.0572 & -- & 0.2831 & 0.6345 & -- & -- & -- & -- \\
RISE-256 & 915 & 6304 & 0.1582 & -- & 0.2962 & 0.6701 & -- & -- & -- & -- \\
ADAPT-TF & 915 & 6304 & 0.0659 & -- & 0.2842 & 0.6260 & -- & -- & -- & -- \\
EigenCAM & 915 & 6304 & 0.1197 & -- & 0.3213 & 0.6761 & -- & -- & -- & -- \\
HiResCAM & 915 & 6304 & 0.1766 & -- & 0.3126 & 0.6955 & -- & -- & -- & -- \\
LayerCAM & 915 & 6304 & 0.2472 & -- & 0.3344 & 0.7296 & -- & -- & -- & -- \\
\midrule
\rowcolor{gray!15}\multicolumn{11}{l}{\textit{Vision-encoder contribution methods}} \\
DINOV3 & 915 & 6304 & 0.3088 & 0.2190 & 0.4667 & 0.7931 & 0.5131 & 0.1536 & 0.2440 & 0.1647 \\
CLIP & 915 & 6304 & 0.3214 & 0.2159 & 0.4710 & 0.8022 & 0.4856 & 0.1625 & 0.2482 & 0.1358 \\
SigLIP2 & 915 & 6304 & \underline{0.3406} & \underline{0.2920} & \underline{0.5104} & \underline{0.8155} & \underline{0.5512} & \underline{0.1394} & \underline{0.2179} & \underline{0.3340} \\
\rowcolor{orange!25!yellow!30}\textbf{IECLA (ours)} & 915 & 6304 & \textbf{0.4804} & \textbf{0.4519} & \textbf{0.5530} & \textbf{0.8636} & \textbf{0.7197} & \textbf{0.1034} & \textbf{0.2025} & \textbf{0.4247} \\
\bottomrule
\end{tabular}%
\end{table}

\begin{table}[H]
\appendixtablesetup
\caption{Cross-dataset attribution results on FI (1,544 images, 9,966 masks).}
\label{tab:cross-fi}
\begin{tabular}{lrrcccccccc}
\toprule
Method & Images & Masks & Imp. $\rho\uparrow$ & Signed $\rho\uparrow$ & Top-1$\uparrow$ & NDCG@3$\uparrow$ & Macro-F1$\uparrow$ & MAE$\downarrow$ & RMSE$\downarrow$ & $R^2\uparrow$ \\
\midrule
\rowcolor{gray!15}\multicolumn{11}{l}{\textit{Heatmap attribution methods}} \\
Ablation-CAM & 1544 & 9966 & 0.0103 & -- & 0.3394 & 0.6174 & -- & -- & -- & -- \\
Grad-CAM & 1544 & 9966 & 0.0023 & -- & 0.3342 & 0.6281 & -- & -- & -- & -- \\
Direct attention & 1544 & 9966 & 0.0418 & -- & 0.3154 & 0.6378 & -- & -- & -- & -- \\
Grad-CAM++ & 1544 & 9966 & 0.0599 & -- & 0.3549 & 0.6463 & -- & -- & -- & -- \\
RISE-256 & 1544 & 9966 & 0.0925 & -- & 0.3472 & 0.6604 & -- & -- & -- & -- \\
ADAPT-TF & 1544 & 9966 & 0.0182 & -- & 0.3187 & 0.6253 & -- & -- & -- & -- \\
EigenCAM & 1544 & 9966 & 0.0972 & -- & 0.3659 & 0.6806 & -- & -- & -- & -- \\
HiResCAM & 1544 & 9966 & 0.1519 & -- & 0.3763 & 0.6963 & -- & -- & -- & -- \\
LayerCAM & 1544 & 9966 & 0.1874 & -- & 0.3789 & 0.6995 & -- & -- & -- & -- \\
\midrule
\rowcolor{gray!15}\multicolumn{11}{l}{\textit{Vision-encoder contribution methods}} \\
DINOV3 & 1544 & 9966 & 0.2733 & 0.1345 & 0.4644 & 0.7826 & 0.5017 & 0.1505 & 0.2486 & 0.1187 \\
CLIP & 1544 & 9966 & 0.2875 & 0.1157 & 0.4903 & 0.7887 & 0.4662 & 0.1588 & 0.2514 & 0.0983 \\
SigLIP2 & 1544 & 9966 & \underline{0.3002} & \underline{0.1869} & \underline{0.4961} & \underline{0.7956} & \underline{0.5193} & \underline{0.1373} & \underline{0.2191} & \underline{0.3155} \\
\rowcolor{orange!25!yellow!30}\textbf{IECLA (ours)} & 1544 & 9966 & \textbf{0.4770} & \textbf{0.4715} & \textbf{0.6101} & \textbf{0.8825} & \textbf{0.7884} & \textbf{0.0746} & \textbf{0.1675} & \textbf{0.5999} \\
\bottomrule
\end{tabular}%
\end{table}

\begin{table}[H]
\appendixtablesetup
\caption{Cross-dataset attribution results on Emotion6 (226 images, 1,335 masks).}
\label{tab:cross-emotion6}
\begin{tabular}{lrrcccccccc}
\toprule
Method & Images & Masks & Imp. $\rho\uparrow$ & Signed $\rho\uparrow$ & Top-1$\uparrow$ & NDCG@3$\uparrow$ & Macro-F1$\uparrow$ & MAE$\downarrow$ & RMSE$\downarrow$ & $R^2\uparrow$ \\
\midrule
\rowcolor{gray!15}\multicolumn{11}{l}{\textit{Heatmap attribution methods}} \\
Ablation-CAM & 226 & 1335 & 0.0104 & -- & 0.3584 & 0.6734 & -- & -- & -- & -- \\
Grad-CAM & 226 & 1335 & 0.0531 & -- & 0.3496 & 0.6925 & -- & -- & -- & -- \\
Direct attention & 226 & 1335 & -0.1231 & -- & 0.2788 & 0.6421 & -- & -- & -- & -- \\
Grad-CAM++ & 226 & 1335 & 0.1207 & -- & 0.3274 & 0.6914 & -- & -- & -- & -- \\
RISE-256 & 226 & 1335 & 0.0549 & -- & 0.3673 & 0.6960 & -- & -- & -- & -- \\
ADAPT-TF & 226 & 1335 & 0.0520 & -- & 0.3496 & 0.6353 & -- & -- & -- & -- \\
EigenCAM & 226 & 1335 & 0.1091 & -- & 0.3894 & 0.7102 & -- & -- & -- & -- \\
HiResCAM & 226 & 1335 & 0.1617 & -- & 0.3496 & 0.7149 & -- & -- & -- & -- \\
LayerCAM & 226 & 1335 & 0.1476 & -- & 0.3673 & 0.7192 & -- & -- & -- & -- \\
\midrule
\rowcolor{gray!15}\multicolumn{11}{l}{\textit{Vision-encoder contribution methods}} \\
CLIP & 226 & 1335 & 0.2581 & 0.1617 & 0.4735 & 0.8059 & 0.4561 & 0.2005 & 0.3051 & 0.0892 \\
DINOV3 & 226 & 1335 & 0.3458 & 0.2180 & 0.4779 & 0.8366 & 0.4731 & 0.1950 & 0.3122 & 0.0463 \\
SigLIP2 & 226 & 1335 & \underline{0.3488} & \underline{0.2283} & \underline{0.5398} & \underline{0.8387} & \underline{0.4982} & \underline{0.1775} & \underline{0.2764} & \underline{0.2523} \\
\rowcolor{orange!25!yellow!30}\textbf{IECLA (ours)} & 226 & 1335 & \textbf{0.5188} & \textbf{0.4822} & \textbf{0.6106} & \textbf{0.9002} & \textbf{0.7400} & \textbf{0.1107} & \textbf{0.2324} & \textbf{0.4716} \\
\bottomrule
\end{tabular}%
\end{table}

\begin{table}[H]
\appendixtablesetup
\caption{Cross-dataset attribution results on EmoArt (73 images, 533 masks).}
\label{tab:cross-emoart}
\begin{tabular}{lrrcccccccc}
\toprule
Method & Images & Masks & Imp. $\rho\uparrow$ & Signed $\rho\uparrow$ & Top-1$\uparrow$ & NDCG@3$\uparrow$ & Macro-F1$\uparrow$ & MAE$\downarrow$ & RMSE$\downarrow$ & $R^2\uparrow$ \\
\midrule
\rowcolor{gray!15}\multicolumn{11}{l}{\textit{Heatmap attribution methods}} \\
Ablation-CAM & 73 & 533 & -0.1059 & -- & 0.1781 & 0.5470 & -- & -- & -- & -- \\
Grad-CAM & 73 & 533 & 0.0319 & -- & 0.2055 & 0.6264 & -- & -- & -- & -- \\
Direct attention & 73 & 533 & -0.0193 & -- & 0.1507 & 0.5811 & -- & -- & -- & -- \\
Grad-CAM++ & 73 & 533 & 0.0793 & -- & 0.1918 & 0.6302 & -- & -- & -- & -- \\
RISE-256 & 73 & 533 & 0.1492 & -- & 0.2192 & 0.6854 & -- & -- & -- & -- \\
ADAPT-TF & 73 & 533 & -0.0326 & -- & 0.1370 & 0.5921 & -- & -- & -- & -- \\
EigenCAM & 73 & 533 & 0.0586 & -- & 0.2055 & 0.6684 & -- & -- & -- & -- \\
HiResCAM & 73 & 533 & 0.1681 & -- & 0.2740 & 0.6730 & -- & -- & -- & -- \\
LayerCAM & 73 & 533 & 0.0541 & -- & 0.1781 & 0.6315 & -- & -- & -- & -- \\
\midrule
\rowcolor{gray!15}\multicolumn{11}{l}{\textit{Vision-encoder contribution methods}} \\
CLIP & 73 & 533 & 0.3520 & 0.3301 & 0.3562 & 0.7994 & 0.4815 & 0.1946 & 0.2780 & 0.1517 \\
DINOV3 & 73 & 533 & 0.3616 & 0.2625 & 0.3699 & \underline{0.8284} & \underline{0.5296} & 0.1798 & 0.2729 & 0.1827 \\
SigLIP2 & 73 & 533 & \underline{0.3858} & \underline{0.3425} & \textbf{0.4384} & 0.7920 & 0.5253 & \underline{0.1768} & \underline{0.2597} & \underline{0.2600} \\
\rowcolor{orange!25!yellow!30}\textbf{IECLA (ours)} & 73 & 533 & \textbf{0.5028} & \textbf{0.5103} & \textbf{0.4384} & \textbf{0.8488} & \textbf{0.7069} & \textbf{0.1208} & \textbf{0.2176} & \textbf{0.4801} \\
\bottomrule
\end{tabular}%
\end{table}

\begin{table}[H]
\appendixtablesetup
\caption{Pooled cross-dataset attribution results under the shared counterfactual teacher. Ranking metrics are image-weighted, whereas Sign Macro-F1 and regression metrics are pooled over valid regions. IECLA leads all reported metrics.}
\label{tab:cross-dataset-attribution-summary}
\centering
\setlength{\tabcolsep}{4pt}
\renewcommand{\arraystretch}{0.90}
\begin{tabular}{lcccccccc}
\toprule
Method & Imp. $\rho\uparrow$ & Signed $\rho\uparrow$ & Top-1$\uparrow$ & NDCG@3$\uparrow$ & Macro-F1$\uparrow$ & MAE$\downarrow$ & RMSE$\downarrow$ & $R^2\uparrow$ \\
\midrule
\rowcolor{gray!15}\multicolumn{9}{l}{\textit{Heatmap attribution methods}} \\
Ablation-CAM~\citeyearpar{desai2020ablationcam} & -0.0137 & -- & 0.2869 & 0.5909 & -- & -- & -- & -- \\
Grad-CAM~\citeyearpar{selvaraju2017gradcam} & 0.0212 & -- & 0.2967 & 0.6214 & -- & -- & -- & -- \\
Direct attention & 0.0304 & -- & 0.2777 & 0.6198 & -- & -- & -- & -- \\
Grad-CAM++~\citeyearpar{chattopadhay2018gradcampp} & 0.0655 & -- & 0.3053 & 0.6326 & -- & -- & -- & -- \\
RISE-256~\citeyearpar{petsiuk2018rise} & 0.1229 & -- & 0.3136 & 0.6548 & -- & -- & -- & -- \\
ADAPT-TF~\citeyearpar{yao2026adapt} & 0.0279 & -- & 0.2843 & 0.6050 & -- & -- & -- & -- \\
EigenCAM~\citeyearpar{muhammad2020eigencam} & 0.1056 & -- & 0.3315 & 0.6656 & -- & -- & -- & -- \\
HiResCAM~\citeyearpar{draelos2020use} & 0.1654 & -- & 0.3332 & 0.6869 & -- & -- & -- & -- \\
LayerCAM~\citeyearpar{jiang2021layercam} & 0.2033 & -- & 0.3445 & 0.6995 & -- & -- & -- & -- \\
\midrule
\rowcolor{gray!15}\multicolumn{9}{l}{\textit{Vision-encoder contribution methods}} \\
CLIP~\citeyearpar{radford2021clip} & 0.3083 & 0.1659 & 0.4727 & 0.7873 & 0.4728 & 0.1637 & 0.2534 & 0.0987 \\
DINOV3~\citeyearpar{simeoni2025dinov3} & 0.3101 & 0.1773 & 0.4567 & 0.7858 & 0.5031 & 0.1536 & 0.2501 & 0.1219 \\
SigLIP2~\citeyearpar{tschannen2025siglip2} & \underline{0.3351} & \underline{0.2423} & \underline{0.4958} & \underline{0.8010} & \underline{0.5331} & \underline{0.1396} & \underline{0.2211} & \underline{0.3137} \\
\rowcolor{orange!25!yellow!30}\textbf{IECLA (ours)} & \textbf{0.5006} & \textbf{0.4830} & \textbf{0.5843} & \textbf{0.8748} & \textbf{0.7600} & \textbf{0.0858} & \textbf{0.1806} & \textbf{0.5422} \\
\bottomrule
\end{tabular}%
\end{table}

The pooled table is not an equal-weight average of the five datasets: its ranking columns weight datasets by the number of valid images, while its sign and regression columns pool all valid regions.

Across all five datasets, IECLA maintains Importance and Signed Spearman correlations above 0.47 and 0.45, respectively, and ranks first on every signed, list-level, and regression metric. The pooled advantage in Table~\ref{tab:cross-dataset-attribution-summary} is therefore not driven by a single source. It persists on FI, the largest external set, and on EmoArt, the smallest and most stylistically distinct set, where IECLA ties SigLIP2 on Top-1 but retains the lead in region ordering, signed agreement, and contribution regression.

The method families exhibit a consistent hierarchy across datasets. Vision-encoder contribution predictors generally improve the available ranking metrics over conventional heatmap attribution methods, indicating that semantic representations better match the teacher-defined regional evidence. IECLA further improves upon the strongest vision-encoder baseline in both signed criteria and contribution-error measures. This joint gain shows that its transfer extends beyond selecting the most relevant region to estimating whether each region supports or opposes the source emotion and by how much. Together, the results establish consistent agreement with the frozen counterfactual teacher across natural, affective, and artistic image distributions.

\section{Reference Validation}
\label{app:reference-validation}

This appendix collects the complete attribution validation associated with Section~\ref{sec:reference-validation}. The three VLM references are evaluated on the same 289 images and the same 1,032 aggregated SAM regions. Method prediction maps, heatmaps, and aggregation rules remain fixed across references, and only the external contribution labels vary. No VLM output is used for training. The two Qwen references provide within-family checks, while GPT Terra supplies a reference from a different model family. Tables~\ref{tab:vlm-qwen-chat}--\ref{tab:vlm-gpt-terra} report image-macro results for each reference, and Table~\ref{tab:vlm-average} gives their unweighted mean before the independent Human Review comparison.

\subsection{VLM Reference}
\label{app:vlm-reference-details}

Qwen Chat and Qwen Reasoner both use Qwen3.8 through the official Qwen API, under direct-response and reasoning configurations, respectively. GPT Terra uses GPT-5.6-Terra. These models serve only as external attribution references and do not participate in contribution-predictor training, checkpoint selection, or editing-agent planning.

\begin{table}[H]
\centering
\begin{minipage}[t]{0.49\textwidth}
\vspace{0pt}
\appendixtablesetup
\setlength{\tabcolsep}{0.8pt}
\caption{Attribution agreement with the Qwen Chat reference. IECLA leads all five metrics.}
\label{tab:vlm-qwen-chat}
\begin{tabular}{lccccc}
\toprule
Method & Imp. $\rho\uparrow$ & Signed $\rho\uparrow$ & Top-1$\uparrow$ & NDCG@3$\uparrow$ & Pointing$\uparrow$ \\
\midrule
\rowcolor{gray!15}\multicolumn{6}{l}{\textit{Heatmap attribution methods}} \\
Ablation-CAM & -0.2294 & -- & 0.4014 & 0.7097 & 0.2388 \\
Grad-CAM & -0.0161 & -- & 0.4775 & 0.7731 & 0.2768 \\
Direct attention & 0.0315 & -- & 0.4810 & 0.7790 & 0.2111 \\
Grad-CAM++ & 0.0843 & -- & 0.5363 & 0.7933 & 0.2630 \\
EigenCAM & 0.0538 & -- & 0.5156 & 0.7862 & 0.2699 \\
HiResCAM & 0.1158 & -- & 0.5571 & 0.7929 & 0.2837 \\
RISE-256 & 0.0748 & -- & 0.5017 & 0.7738 & 0.4394 \\
LayerCAM & 0.1835 & -- & 0.5675 & \underline{0.8144} & 0.2699 \\
ADAPT-TF & \underline{0.1890} & -- & 0.5606 & 0.8081 & 0.2941 \\
\midrule
\rowcolor{gray!15}\multicolumn{6}{l}{\textit{Vision-encoder contribution methods}} \\
CLIP & 0.1107 & 0.0407 & 0.5744 & 0.7936 & 0.6678 \\
SigLIP2 & 0.1527 & 0.1280 & \underline{0.6159} & 0.8083 & \underline{0.6851} \\
DINOV3 & 0.1507 & \underline{0.1688} & 0.6125 & 0.8083 & 0.6782 \\
\rowcolor{orange!25!yellow!30}\textbf{IECLA (ours)} & \textbf{0.6187} & \textbf{0.4868} & \textbf{0.8720} & \textbf{0.9331} & \textbf{0.8685} \\
\bottomrule
\end{tabular}%
\end{minipage}\hfill%
\begin{minipage}[t]{0.49\textwidth}
\vspace{0pt}
\appendixtablesetup
\setlength{\tabcolsep}{0.8pt}
\caption{Qwen Reasoner reference results. IECLA retains its lead across all five metrics.}
\label{tab:vlm-qwen-reasoner}
\begin{tabular}{lccccc}
\toprule
Method & Imp. $\rho\uparrow$ & Signed $\rho\uparrow$ & Top-1$\uparrow$ & NDCG@3$\uparrow$ & Pointing$\uparrow$ \\
\midrule
\rowcolor{gray!15}\multicolumn{6}{l}{\textit{Heatmap attribution methods}} \\
Ablation-CAM & -0.2647 & -- & 0.3772 & 0.6826 & 0.2249 \\
Grad-CAM & -0.0222 & -- & 0.4913 & 0.7578 & 0.2630 \\
Direct attention & 0.0340 & -- & 0.4844 & 0.7722 & 0.2007 \\
Grad-CAM++ & 0.0988 & -- & 0.5467 & 0.7746 & 0.2491 \\
EigenCAM & 0.0193 & -- & 0.5052 & 0.7712 & 0.2526 \\
HiResCAM & 0.1695 & -- & 0.5848 & 0.7942 & 0.2699 \\
RISE-256 & 0.1335 & -- & 0.5225 & 0.7734 & 0.4187 \\
LayerCAM & \underline{0.2348} & -- & 0.5813 & \underline{0.8158} & 0.2457 \\
ADAPT-TF & 0.2291 & -- & 0.5779 & 0.8021 & 0.2907 \\
\midrule
\rowcolor{gray!15}\multicolumn{6}{l}{\textit{Vision-encoder contribution methods}} \\
CLIP & 0.1088 & 0.0501 & 0.5640 & 0.7827 & 0.6540 \\
SigLIP2 & 0.1142 & 0.1370 & \underline{0.5952} & 0.7873 & \underline{0.6574} \\
DINOV3 & 0.1163 & \underline{0.1741} & 0.5882 & 0.7831 & \underline{0.6574} \\
\rowcolor{orange!25!yellow!30}\textbf{IECLA (ours)} & \textbf{0.5998} & \textbf{0.5108} & \textbf{0.8789} & \textbf{0.9313} & \textbf{0.8616} \\
\bottomrule
\end{tabular}%
\end{minipage}\par\vspace{6pt}
\begin{minipage}[t]{0.49\textwidth}
\vspace{0pt}
\appendixtablesetup
\setlength{\tabcolsep}{0.8pt}

\caption{GPT Terra reference results. IECLA retains its lead under a different VLM family.}
\label{tab:vlm-gpt-terra}
\begin{tabular}{lccccc}
\toprule
Method & Imp. $\rho\uparrow$ & Signed $\rho\uparrow$ & Top-1$\uparrow$ & NDCG@3$\uparrow$ & Pointing$\uparrow$ \\
\midrule
\rowcolor{gray!15}\multicolumn{6}{l}{\textit{Heatmap attribution methods}} \\
Ablation-CAM & -0.1633 & -- & 0.4048 & 0.7231 & 0.2388 \\
Grad-CAM & 0.0698 & -- & 0.5121 & 0.7959 & 0.2803 \\
Direct attention & 0.0077 & -- & 0.4221 & 0.7664 & 0.2215 \\
Grad-CAM++ & 0.0863 & -- & 0.5121 & 0.7971 & 0.2803 \\
EigenCAM & 0.0460 & -- & 0.4706 & 0.7851 & 0.2803 \\
HiResCAM & 0.1044 & -- & 0.5363 & 0.7960 & 0.2941 \\
RISE-256 & -0.0277 & -- & 0.4256 & 0.7524 & 0.4394 \\
LayerCAM & 0.1517 & -- & 0.5225 & 0.8119 & 0.2837 \\
ADAPT-TF & 0.0418 & -- & 0.4844 & 0.7832 & 0.2699 \\
\midrule
\rowcolor{gray!15}\multicolumn{6}{l}{\textit{Vision-encoder contribution methods}} \\
CLIP & 0.1850 & 0.0231 & 0.5882 & 0.8121 & 0.6609 \\
SigLIP2 & 0.2138 & \underline{0.0705} & 0.6055 & \underline{0.8275} & \underline{0.6644} \\
DINOV3 & \underline{0.2252} & 0.0626 & \underline{0.6159} & 0.8261 & 0.6574 \\
\rowcolor{orange!25!yellow!30}\textbf{IECLA (ours)} & \textbf{0.6165} & \textbf{0.5324} & \textbf{0.8789} & \textbf{0.9323} & \textbf{0.8547} \\
\bottomrule
\end{tabular}%
\end{minipage}\hfill%
\begin{minipage}[t]{0.49\textwidth}
\vspace{0pt}
\appendixtablesetup
\setlength{\tabcolsep}{0.8pt}
\caption{Mean agreement across VLM references. IECLA leads every metric.}
\label{tab:vlm-average}
\begin{tabular}{lccccc}
\toprule
Method & Imp. $\rho\uparrow$ & Signed $\rho\uparrow$ & Top-1$\uparrow$ & NDCG@3$\uparrow$ & Pointing$\uparrow$ \\
\midrule
\rowcolor{gray!15}\multicolumn{6}{l}{\textit{Heatmap attribution methods}} \\
Ablation-CAM & -0.2191 & -- & 0.3945 & 0.7051 & 0.2342 \\
Grad-CAM & 0.0105 & -- & 0.4936 & 0.7756 & 0.2734 \\
Direct attention & 0.0244 & -- & 0.4625 & 0.7725 & 0.2111 \\
Grad-CAM++ & 0.0898 & -- & 0.5317 & 0.7883 & 0.2641 \\
EigenCAM & 0.0397 & -- & 0.4971 & 0.7808 & 0.2676 \\
HiResCAM & 0.1299 & -- & 0.5594 & 0.7944 & 0.2826 \\
RISE-256 & 0.0602 & -- & 0.4833 & 0.7665 & 0.4325 \\
LayerCAM & \underline{0.1900} & -- & 0.5571 & \underline{0.8140} & 0.2664 \\
ADAPT-TF & 0.1533 & -- & 0.5410 & 0.7978 & 0.2849 \\
\midrule
\rowcolor{gray!15}\multicolumn{6}{l}{\textit{Vision-encoder contribution methods}} \\
CLIP & 0.1348 & 0.0380 & 0.5755 & 0.7961 & 0.6609 \\
SigLIP2 & 0.1602 & 0.1118 & \underline{0.6055} & 0.8077 & \underline{0.6690} \\
DINOV3 & 0.1641 & \underline{0.1352} & \underline{0.6055} & 0.8058 & 0.6643 \\
\rowcolor{orange!25!yellow!30}\textbf{IECLA (ours)} & \textbf{0.6117} & \textbf{0.5100} & \textbf{0.8766} & \textbf{0.9322} & \textbf{0.8616} \\
\bottomrule
\end{tabular}%
\end{minipage}
\end{table}

The reference-specific results expose a stability pattern that is hidden by the average alone. Across Qwen Chat, Qwen Reasoner, and GPT Terra, IECLA's Top-1, NDCG@3, and Pointing scores vary by at most 0.014, although the strongest competing method changes with the metric and reference. Importance and Signed Spearman show larger reference-dependent shifts, yet IECLA remains ahead of the competing methods. The result is therefore supported by agreement across references rather than an isolated match to one evaluator.

The baseline profiles also separate localization from signed attribution. Heatmap methods can obtain competitive NDCG@3 or Pointing on individual references, but their nonnegative outputs cannot represent opposing evidence. Vision-encoder contribution predictors add signed estimates and improve region localization, while their signed agreement remains substantially below IECLA. The simultaneous gain in ranking, sign, and Pointing indicates that the contribution predictor aligns these complementary properties instead of optimizing a single localization criterion.

\FloatBarrier
\subsection{Human Reference}
\label{app:human-reference-details}
Human Review starts from 320 emotion-balanced images; fixed usability filtering and human-region merging define the reported evaluation set. Forty annotators complete disjoint batches, with each item assigned to one annotator. All methods use the same SAM candidate masks and region mapping. We compute the ranking and localization metrics in Table~\ref{tab:human-attribution} per image and average over images. Signed metrics are reported only for methods that produce signed contributions.

\begin{table}[H]
\appendixtablesetup
\caption{Attribution agreement with independent Human Review annotations. IECLA leads in region ranking, signed agreement, and localization.}
\label{tab:human-attribution}
\begin{tabular}{lccccc}
\toprule
Method & Imp. $\rho\uparrow$ & Signed $\rho\uparrow$ & Top-1$\uparrow$ & NDCG@3$\uparrow$ & Pointing$\uparrow$ \\
\midrule
\rowcolor{gray!15}\multicolumn{6}{l}{\textit{Heatmap attribution methods}} \\
Ablation-CAM & -0.2207 & -- & 0.5502 & 0.7809 & 0.3218 \\
Grad-CAM & -0.0450 & -- & 0.5952 & 0.8159 & 0.3426 \\
Direct attention & -0.0294 & -- & 0.5744 & 0.8214 & 0.2595 \\
Grad-CAM++ & 0.0115 & -- & 0.6332 & 0.8272 & 0.3391 \\
EigenCAM & 0.0491 & -- & 0.6298 & 0.8321 & 0.3495 \\
HiResCAM & 0.0707 & -- & 0.6678 & 0.8474 & 0.3495 \\
RISE-256 & 0.1264 & -- & 0.6574 & 0.8397 & 0.4706 \\
LayerCAM & 0.1526 & -- & 0.6886 & 0.8572 & 0.3460 \\
ADAPT-TF & \underline{0.1942} & -- & 0.6713 & 0.8464 & 0.3529 \\
\midrule
\rowcolor{gray!15}\multicolumn{6}{l}{\textit{Vision-encoder contribution methods}} \\
CLIP & 0.0726 & 0.0547 & 0.6851 & 0.8492 & 0.7543 \\
DINOV3 & 0.1373 & \underline{0.1164} & \underline{0.7093} & \underline{0.8662} & \underline{0.7716} \\
SigLIP2 & 0.1412 & 0.0921 & \underline{0.7093} & 0.8619 & 0.7509 \\
\rowcolor{orange!25!yellow!30}\textbf{IECLA (ours)} & \textbf{0.5001} & \textbf{0.4525} & \textbf{0.8685} & \textbf{0.9315} & \textbf{0.8478} \\
\bottomrule
\end{tabular}%
\end{table}

Comparing Table~\ref{tab:human-attribution} with the VLM average in Table~\ref{tab:vlm-average} reveals which properties transfer most closely to human judgments. IECLA's NDCG@3 is nearly unchanged, while Top-1 and Pointing decrease only modestly; the larger reductions occur in Importance and Signed Spearman correlation. Thus, the leading region and top-weighted ordering are more stable across reference types than the full ordering of regional contributions. IECLA remains first throughout, whereas the strongest alternative changes between ADAPT-TF, DINOV3, and SigLIP2 depending on the criterion.

\paragraph{Signed contribution classification.}
Signed direction is evaluated independently on the same Human Review annotations. The protocol starts from 320 images and 1,319 annotated regions; filtering leaves 289 images and 1,032 human regions. When a human region overlaps several SAM masks, source-mask scores are aggregated with source-mask-area weights. The Sign Macro-F1 values in Table~\ref{tab:human-sign-macro-f1} use image-macro aggregation with 10,000 image-level bootstrap resamples, and signed methods use the fixed threshold $\tau=0.1$. Nonnegative attention and saliency methods are omitted because they do not produce signed outputs.

IECLA reaches a Sign Macro-F1 of 0.3601, more than twice the strongest competing score of 0.1765. Because this classification uses independent Human Review annotations and a fixed sign threshold, the result complements the continuous signed-correlation analysis. IECLA aligns more closely with both regional ordering and the discrete distinction between evidence that supports or opposes the perceived source emotion.

This comparison tests the usefulness of the calibrated sign for distinguishing supporting and opposing perceptual evidence; it does not establish an absolute causal polarity for every candidate region.

\FloatBarrier

\section{User Study}
\label{app:user-study-details}

This appendix provides the complete protocol and detailed results for the blinded User Study in Section~\ref{sec:user-study}. The study has two complementary components: a heatmap evaluation of whether the attributed regions are accurate and easy to interpret, and a paired editing evaluation of whether the same localized evidence improves target-emotion expression while preserving naturalness. We randomize candidate order independently within each task.

\paragraph{Protocol and scoring.}
Forty independent reviewers score the anonymized candidates on a five-level scale from $-2$ to $2$. Heatmap candidates are evaluated by source-evidence accuracy and focus clarity; editing pairs are evaluated by target expression and naturalness. After decoding the blinded assignments, we aggregate reviewer scores. Table~\ref{tab:user-agreement} summarizes the evaluation units and the corresponding ordinal reliability coefficients.

\begin{table}[H]
\centering
\begin{minipage}[t]{0.48\linewidth}
\appendixtablesetup
\captionof{table}{Signed contribution classification against Human Review.}
\label{tab:human-sign-macro-f1}
\begin{tabular}{lc}
\toprule
Method & Sign Macro-F1$\uparrow$ \\
\midrule
DINOV3 & 0.1530 \\
CLIP & 0.1711 \\
SigLIP2 & \underline{0.1765} \\
\rowcolor{orange!25!yellow!30}\textbf{IECLA (ours)} & \textbf{0.3601} \\
\bottomrule
\end{tabular}
\end{minipage}\hfill
\hspace{-6pt}\begin{minipage}[t]{0.48\linewidth}
\appendixtablesetup
\captionof{table}{Ordinal inter-rater reliability in the blinded User Study. Each evaluation unit is rated by all 40 reviewers.}
\label{tab:user-agreement}
\begin{tabular}{lrrc}
\toprule
Component & Tasks & Evaluation units & Krippendorff's $\alpha$ \\
\midrule
Heatmap & 20 & 520 & 0.8392 \\
Editing & 120 paired & 480 & 0.7364 \\
Overall & -- & 1,000 & -- \\
\bottomrule
\end{tabular}
\end{minipage}
\end{table}

The two components exhibit different levels of ordinal inter-rater reliability. Heatmap ratings reach a Krippendorff's $\alpha$ of 0.8392, while the paired editing ratings reach 0.7364. These coefficients are computed from the unaggregated ratings of all 40 reviewers and measure reviewer consistency rather than method superiority. Reviewer means are used only for the subsequent task-level comparisons, so the reliability results support the interpretation of the study without replacing the task-level tests.

\paragraph{Heatmap evaluation.}
Each heatmap task fixes the source image and target emotion, then presents anonymized candidate maps generated under the same candidate-region protocol. The two criteria are averaged per candidate and then across the 20 tasks. The complete comparison is reproduced in Table~\ref{tab:user-heatmap-details}. Its Friedman statistic and Kendall's $W$ indicate that the observed ordering is descriptive because the repeated-measures test does not reach significance.

\begin{wraptable}{r}{0.46\textwidth}
\vspace{-8pt}
\appendixtablesetup
\captionof{table}{Detailed heatmap User Study results. Mean score averages source-evidence accuracy and focus clarity; top-score counts include task-level ties among 20 tasks.}
\label{tab:user-heatmap-details}
\begin{tabular}{@{}lcc@{}}
\toprule
Method & Mean score$\uparrow$ & Top score$\uparrow$ \\
\midrule
LayerCAM & 0.775 & 10 \\
CLIP & 0.925 & 12 \\
HiResCAM & 0.950 & 13 \\
ADAPT-TF & 0.950 & 13 \\
DINOV3 & 0.975 & 13 \\
Grad-CAM++ & 0.975 & 14 \\
EigenCAM & 0.975 & 13 \\
RISE-256 & \underline{1.025} & \underline15 \\
SigLIP2 & \underline{1.025} & 14 \\
Direct attention & \underline{1.025} & 14 \\
Grad-CAM & \underline{1.025} & 15 \\
Ablation-CAM & \underline{1.025} & \underline{16} \\
\rowcolor{orange!25!yellow!30}\textbf{IECLA (ours)} & \textbf{1.075} & \textbf{18} \\
\bottomrule
\end{tabular}
\vspace{-6pt}
\end{wraptable}

The heatmap scores are tightly clustered: most methods lie between 0.925 and 1.025, and task-level ties are frequent. Within this compressed range, IECLA attains the highest mean and appears among the top-scoring candidates on 18 of 20 tasks.

\paragraph{Editing evaluation.}
For each editor, the same 20 source--target jobs are rendered with the native configuration and with IECLA-guided localized evidence. Each pair fixes the source image, target emotion, editor, and corresponding generation configuration, while the A/B order is randomized independently within each task. The table reports reviewer means and within-editor differences. Mean ratings for both target expression and naturalness increase across all six editors, consistent with the main-paper comparison in Section~\ref{sec:user-study}. Reviewer means are computed within each evaluation unit before the paired task-level comparison, while ordinal reliability is reported separately for the heatmap and editing components in Table~\ref{tab:user-agreement}.

\begin{table}[!ht]
\appendixtablesetup
\caption{Paired editing ratings across six backbones. IECLA guidance improves both target expression and naturalness for every editor.}
\label{tab:user-editing}
\begin{tabular}{lcccccc}
\toprule
& \multicolumn{3}{c}{\cellcolor{gray!15}Target expression $\uparrow$} & \multicolumn{3}{c}{\cellcolor{gray!15}Naturalness $\uparrow$} \\
\cmidrule(lr){2-4} \cmidrule(lr){5-7}
Editor & Native & IECLA-guided & $\Delta$ & Native & IECLA-guided & $\Delta$ \\
\midrule
Qwen-Edit & 0.0625 & 1.325 & \cellcolor{orange!25!yellow!30}\textbf{+1.2625} & 0.375 & 1.025 & \cellcolor{orange!25!yellow!30}\textbf{+0.650} \\
HiDream & 0.200 & 1.400 & \cellcolor{orange!25!yellow!30}\textbf{+1.200} & 0.425 & 0.675 & \cellcolor{orange!25!yellow!30}\textbf{+0.250} \\
InstructPix2Pix & 0.225 & 0.400 & \cellcolor{orange!25!yellow!30}\textbf{+0.175} & 0.500 & 0.575 & \cellcolor{orange!25!yellow!30}\textbf{+0.075} \\
FLUX.2 & -0.350 & 0.150 & \cellcolor{orange!25!yellow!30}\textbf{+0.500} & 1.050 & 1.425 & \cellcolor{orange!25!yellow!30}\textbf{+0.375} \\
OmniGen2 & 0.650 & 1.450 & \cellcolor{orange!25!yellow!30}\textbf{+0.800} & 0.350 & 0.600 & \cellcolor{orange!25!yellow!30}\textbf{+0.250} \\
LongCat & -0.075 & 0.075 & \cellcolor{orange!25!yellow!30}\textbf{+0.150} & 0.125 & 0.200 & \cellcolor{orange!25!yellow!30}\textbf{+0.075} \\
\bottomrule
\end{tabular}%
\end{table}

The sign of the effect is stable across editors, whereas its magnitude is editor dependent. Qwen-Edit and HiDream show the largest target-expression gains, while Qwen-Edit and FLUX.2 show the largest naturalness gains. OmniGen2 already has the highest native target-expression score yet still improves by 0.800, so the benefit is not limited to correcting editors with weak native emotion control. InstructPix2Pix and LongCat exhibit smaller but positive changes on both criteria, indicating that the response to localized evidence varies across editing architectures.

The paired pattern also clarifies the relation between the two subjective objectives. Stronger target expression does not coincide with a naturalness decrease for any of the six editors in this evaluation; all naturalness differences remain positive. Together with the variation in gain magnitude, these point estimates suggest that localized IECLA evidence is useful across editing architectures, while the realized benefit remains dependent on the underlying editor.

\FloatBarrier
\section{Additional Ablations}
\label{app:loss-ablation}

Table~\ref{tab:loss-ablation} isolates training and editing decisions that are not separated in the main end-to-end ablation. All variants use the same 680-image, 5,440-task evaluation pool, predictor architecture, editor, and metric protocol; only the named component or supervision term is changed.

\begin{table}[H]
\centering
\appendixtablesetup
\setlength{\tabcolsep}{2.4pt}
\renewcommand{\arraystretch}{0.94}
\caption{Additional editing and loss-function ablations. IECLA provides the strongest overall balance between target-emotion control and structural preservation.}
\label{tab:loss-ablation}
\resizebox{\textwidth}{!}{%
\begin{tabular}{lcccccccccc}
\toprule
Variant & Target Success $\uparrow$ & $\Delta p$ $\uparrow$ & Emo-A $\uparrow$ & Emo-S $\uparrow$ & CLIP-I $\uparrow$ & G. LPIPS $\downarrow$ & G. SSIM $\uparrow$ & O. LPIPS $\downarrow$ & O. SSIM $\uparrow$ & O. MAE $\downarrow$ \\
\midrule
\rowcolor{orange!25!yellow!30}\textbf{IECLA} & \textbf{0.3579} & \textbf{0.3344} & \textbf{0.3081} & \textbf{0.2173} & \textbf{0.8527} & \textbf{0.3440} & \textbf{0.6137} & 0.3767 & \textbf{0.3230} & \underline{0.1733} \\
\quad -w/o $\mathcal{L}_{\mathrm{abs}}$ & 0.3412 & 0.3210 & 0.2901 & \underline{0.2150} & 0.8421 & 0.3507 & 0.6000 & \textbf{0.3721} & \underline{0.3185} & 0.1831 \\
\quad -w/o $\mathcal{L}_{\mathrm{signed}}$ & 0.3471 & 0.3226 & 0.2921 & \underline{0.2150} & 0.8415 & 0.3533 & 0.5975 & 0.3751 & 0.3173 & 0.1842 \\
\quad -w/o $\mathcal{L}_{\mathrm{focal}}$ & 0.3390 & 0.3164 & 0.2899 & 0.2110 & 0.8438 & 0.3515 & 0.5996 & \underline{0.3729} & \underline{0.3185} & 0.1834 \\
\quad -w/o $\mathcal{L}_{\mathrm{aux}}$ & 0.3465 & 0.3221 & 0.2914 & 0.2143 & 0.8418 & 0.3534 & 0.5969 & 0.3752 & 0.3169 & 0.1842 \\
\quad -w/o fusion & \underline{0.3542} & 0.3318 & 0.2921 & 0.2139 &
0.8451 & 0.3476 & \underline{0.6073} & 0.3851 & \underline{0.3214} & 0.1746 \\
\quad -w/o template & 0.3518 & 0.3280 & \underline{0.2969} & 0.2131 &
0.8397 & 0.3524 & 0.6035 & 0.3862 & 0.3180 & 0.1793 \\
\quad -w/o negative protection & 0.3504 & \underline{0.3335} & 0.2820 & 0.1980 & 0.8454 & \textbf{0.3440} & 0.6028 & 0.3839 & 0.3167 & 0.1774 \\
Random matched & 0.2739 & 0.2507 & 0.2706 & 0.1658 & \underline{0.8481} & \underline{0.3442} & 0.6051 & 0.3772 & 0.3143 & \textbf{0.1716} \\
\bottomrule
\end{tabular}%
}
\end{table}

The complete IECLA configuration is the reference condition. The ``w/o fusion'' variant removes global--local context fusion from the contribution predictor, testing whether regional evidence benefits from joint local and scene-level representations. The ``w/o template'' variant retains attribution and agent planning but omits the structured editing template, testing whether the template converts evidence into consistent editing instructions. The ``w/o learning'' variant bypasses the contribution predictor and passes the corrected counterfactual contributions directly from attribution to target-aware editing, testing whether learned contribution representations improve over direct evidence transfer. These settings complement the main-paper ``w/o attribution'', ``w/o learning'', and ``w/o agent'' conditions rather than duplicating them.

Two controls target the use of signed evidence during editing. In \textit{w/o negative protection}, positive contribution regions and the affective instruction are unchanged, but the prompt omits preservation constraints derived from negatively contributing regions. This tests whether negative contributions act as preservation evidence rather than merely being unused scores. In \textit{Random matched}, the attributed positive region is replaced by a deterministically sampled candidate satisfying the same matching criteria, while prompt structure and editing strength are retained. This control tests whether the gain comes from selecting emotionally relevant evidence instead of from structured prompting alone. The fixed seed and image identifier make the sampled candidate reproducible.

The remaining rows remove one loss from Equation~\ref{eq:training-objective} at a time while keeping the predictor, data, and downstream editing protocol fixed. ``w/o $\mathcal{L}_{\mathrm{abs}}$'' removes absolute ranking supervision and tests contribution-magnitude ordering; ``w/o $\mathcal{L}_{\mathrm{signed}}$'' removes signed ranking supervision and tests direction-aware ordering; ``w/o $\mathcal{L}_{\mathrm{focal}}$'' removes focal-sign supervision and tests emphasis on difficult or imbalanced sign decisions; and ``w/o $\mathcal{L}_{\mathrm{aux}}$'' removes auxiliary supervision and tests the contribution of the auxiliary affective signals. Bold and underlined entries denote the best and second-best point estimates in each column.

The editing controls separate evidence selection from instruction structure. Removing negative protection lowers all four affective measures and reduces outside-region SSIM, while G. LPIPS remains tied with IECLA at 0.3440. Negative contributions therefore constrain unnecessary changes without simply suppressing the edit. Random matched causes the largest affective drop, reaching 0.2739 Target Success and 0.1658 Emo-S, even though it obtains the lowest O. MAE (0.1716). The combination of weak emotion control and conservative pixel change shows that the selected region, rather than structured prompting alone, determines whether the edit is useful. Removing fusion or the template produces smaller but consistent losses in affective and structural measures, with the template ablation particularly lowering CLIP-I to 0.8397 and increasing O. LPIPS to 0.3862.

The main-paper component ablations expose three distinct roles in the editing pipeline. Removing attribution leaves the agent without region-specific affective evidence, so its planning cannot reliably identify where the target emotion should be expressed or which regions should be preserved. Removing learning retains the attribution signal but bypasses its learned calibration, context integration, and stable magnitude and sign ordering; direct evidence transfer therefore provides a noisier control signal and permits more unintended changes outside the selected region. Removing the agent retains the learned contribution representation, allowing the editor to use localized evidence directly, but loses the structured decomposition of target-emotion actions and preservation constraints that makes the evidence consistently executable. Together, the three conditions separate the roles of evidence construction, evidence calibration, and evidence-to-instruction planning.

The loss ablations reveal a complementary training effect. Removing focal-sign supervision gives the largest affective degradation among the four loss variants, whereas removing either ranking term or auxiliary supervision produces smaller declines across the affective metrics. Individual variants can improve one preservation distance, such as O. LPIPS after removing absolute ranking, but none matches IECLA's joint profile: the complete model is best on all four affective measures, CLIP-I, G. SSIM, and O. SSIM, while tying the best G. LPIPS value and remaining second-best on O. MAE. The results link the full objective to three functions of the framework: ordering relevant evidence, preserving the direction of signed contributions, and maintaining structural fidelity during editing.

\FloatBarrier
\section{Stability Analysis}
\label{app:stability}

Robustness to sampling variation is assessed with Qwen-Edit, using the configuration in Appendix~\ref{app:implementation-details}, on 16 routable images stratified by the eight source emotions, with two images per class. Each image is edited toward the other seven emotions, yielding 112 matched tasks. The source image, target emotion, agent configuration, routing protocol, and SAM regions are fixed. Native receives only the global target-emotion instruction; IECLA-guided editing augments it with at most three human-related SAM regions whose predicted positive contribution exceeds 0.05. Each pair shares its source image, target emotion, seed, and resolution.

Seeds 48141, 48142, and 48143 are evaluated at the main resolution $832\times480$ and the adjacent resolution $1024\times576$. Each resolution contains $112\times2\times3=672$ outputs, or 1,344 in total. Target Success is the frozen EmoVerse classifier's target decision, and Target probability $\Delta$ measures change relative to the source image. Preservation is measured outside the hard union of routed SAM masks. Affective IECLA-guided--Native differences use 10,000 source-level paired bootstrap resamples. Preservation values remain point differences because verified paired intervals were not retained; resolution-level standard deviations are computed across the three seed-level means.

\begin{table}[!ht]
\appendixtablesetup
\caption{Stability across seeds and resolutions. IECLA shows favorable affective and outside-mask preservation differences in all six settings.}
\label{tab:stability}
\begin{tabular}{lcccc}
\toprule
& \multicolumn{2}{>{\columncolor{gray!15}[\dimexpr\tabcolsep-\cmidrulekern\relax][\tabcolsep]}c}{Affective response} & \multicolumn{2}{c}{\cellcolor{gray!15}Outside-mask preservation} \\
\cmidrule(lr){2-3} \cmidrule(l){4-5}
Seed & \shortstack{$\Delta$ Target Success\\{[95\% CI]}} & \shortstack{$\Delta$ Target prob.\\{[95\% CI]}} & \shortstack{Outside-mask\\$\Delta$ LPIPS} & \shortstack{Outside-mask\\$\Delta$ SSIM} \\
\midrule
\multicolumn{5}{l}{\textit{Resolution: $1024\times576$}} \\
48141 & +0.098 [+0.027, +0.179] & +0.098 [+0.034, +0.169] & -0.010 & +0.001 \\
48142 & +0.107 [+0.018, +0.214] & +0.108 [+0.035, +0.199] & -0.018 & +0.002 \\
48143 & +0.098 [+0.036, +0.170] & +0.078 [+0.036, +0.128] & -0.005 & +0.010 \\
\addlinespace
\multicolumn{5}{l}{\textit{Resolution: $832\times480$}} \\
48141 & +0.045 [-0.009, +0.107] & +0.044 [-0.013, +0.107] & -0.011 & +0.005 \\
48142 & +0.054 [+0.009, +0.116] & +0.051 [+0.004, +0.108] & -0.008 & +0.001 \\
48143 & +0.062 [+0.018, +0.107] & +0.063 [+0.019, +0.104] & -0.005 & +0.008 \\
\bottomrule
\end{tabular}%
\end{table}

IECLA-guided editing improves Target Success and target probability for every seed and resolution. At $1024\times576$, all affective intervals exclude zero; the three-seed mean gains are $+0.101\pm0.005$ for Target Success and $+0.095\pm0.015$ for target probability. At $832\times480$, the corresponding gains are smaller at $+0.054\pm0.009$ and $+0.053\pm0.010$; the seed-48141 intervals cross zero, while the other two exclude zero. Outside-mask LPIPS decreases and SSIM increases in all six settings. The repeated directions support sampling stability within each resolution, while their magnitudes indicate sensitivity to output resolution.

The stability evidence is limited to 16 source images, three seeds, two resolutions, and one editor. It supports consistency of the observed direction under this protocol, rather than universal robustness across editors, samplers, resolutions, or image populations.

\section{Teacher Replay and Predictor Comparison}
\label{app:frozen-teacher}

This appendix isolates whether the learned contribution predictor improves on direct replay of the intervention-based supervision that defines its target. The comparison is useful because the replay is an exact counterfactual measurement, whereas the predictor can denoise that signal and generalize it to the reference-validation protocol.

\paragraph{Experimental setup.}
The replay uses the same frozen Phase-1 emotion evaluator $E$ that constructs the corrected counterfactual contributions in Section~\ref{sec:counterfactual-labels}. For each of the 1,032 aggregated candidate regions on the 289-image reference-validation set, we infer the source emotion with $E$, apply the candidate blur intervention, apply its matched reference intervention, and compute the signed contribution with Equation~\ref{eq:counterfactual-contribution}. No predictor is trained or selected for this condition. Candidate masks, reference-region matching, prediction-map rendering, region aggregation, sign threshold, and image-macro metric computation are identical to the reference-validation protocol. The replay is compared with IECLA's learned predictor against Human Review, Qwen Chat, Qwen Reasoner, GPT Terra, and their unweighted VLM mean.

\begin{table}[H]
\centering
\appendixtablesetup
\scriptsize
\setlength{\tabcolsep}{2.0pt}
\renewcommand{\arraystretch}{0.92}
\newcommand{\teacherdelta}[1]{\cellcolor{orange!25!yellow!30}\textbf{#1}}
\caption{Teacher replay versus the IECLA predictor under the reference-validation protocol.}
\label{tab:frozen-teacher-replay}
\makebox[\textwidth][c]{%
\scalebox{0.93}{%
\begin{tabular}{@{}lccc ccc ccc ccc ccc@{}}
\toprule
\rowcolor{gray!15}
& \multicolumn{3}{c}{Imp.~$\rho\uparrow$}
& \multicolumn{3}{c}{Signed~$\rho\uparrow$}
& \multicolumn{3}{c}{Top-1$\uparrow$}
& \multicolumn{3}{c}{Top-3$\uparrow$}
& \multicolumn{3}{c}{Top-5$\uparrow$} \\
\cmidrule(lr){2-4} \cmidrule(lr){5-7} \cmidrule(lr){8-10} \cmidrule(lr){11-13} \cmidrule(lr){14-16}
Reference & Replay & IECLA & $\Delta$ & Replay & IECLA & $\Delta$ & Replay & IECLA & $\Delta$ & Replay & IECLA & $\Delta$ & Replay & IECLA & $\Delta$ \\
\midrule
Human Review & 0.3730 & \textbf{0.5001} & \teacherdelta{+0.1271} & 0.4094 & \textbf{0.4525} & \teacherdelta{+0.0431} & 0.8097 & \textbf{0.8685} & \teacherdelta{+0.0588} & 0.9758 & \textbf{0.9896} & \teacherdelta{+0.0138} & 0.9965 & \textbf{0.9965} & \teacherdelta{+0.0000} \\
Qwen Chat & 0.4522 & \textbf{0.6187} & \teacherdelta{+0.1665} & 0.4522 & \textbf{0.4868} & \teacherdelta{+0.0346} & 0.7751 & \textbf{0.8720} & \teacherdelta{+0.0969} & 0.9792 & \textbf{0.9896} & \teacherdelta{+0.0104} & 0.9931 & \textbf{1.0000} & \teacherdelta{+0.0069} \\
Qwen Reasoner & 0.4330 & \textbf{0.5998} & \teacherdelta{+0.1668} & 0.4195 & \textbf{0.5108} & \teacherdelta{+0.0913} & 0.7785 & \textbf{0.8789} & \teacherdelta{+0.1004} & 0.9723 & \textbf{0.9792} & \teacherdelta{+0.0069} & 0.9931 & \textbf{1.0000} & \teacherdelta{+0.0069} \\
GPT Terra & 0.4477 & \textbf{0.6165} & \teacherdelta{+0.1688} & 0.4275 & \textbf{0.5324} & \teacherdelta{+0.1049} & 0.7612 & \textbf{0.8789} & \teacherdelta{+0.1177} & 0.9654 & \textbf{0.9827} & \teacherdelta{+0.0173} & 0.9931 & \textbf{1.0000} & \teacherdelta{+0.0069} \\
VLM mean & 0.4443 & \textbf{0.6117} & \teacherdelta{+0.1674} & 0.4330 & \textbf{0.5100} & \teacherdelta{+0.0770} & 0.7716 & \textbf{0.8766} & \teacherdelta{+0.1050} & 0.9723 & \textbf{0.9838} & \teacherdelta{+0.0115} & 0.9931 & \textbf{1.0000} & \teacherdelta{+0.0069} \\
\midrule
\rowcolor{gray!15}
& \multicolumn{3}{c}{NDCG@3$\uparrow$}
& \multicolumn{3}{c}{NDCG@5$\uparrow$}
& \multicolumn{3}{c}{Pointing$\uparrow$}
& \multicolumn{3}{c}{Sign Accuracy$\uparrow$}
& \multicolumn{3}{c}{Sign F1$\uparrow$} \\
\cmidrule(lr){2-4} \cmidrule(lr){5-7} \cmidrule(lr){8-10} \cmidrule(lr){11-13} \cmidrule(lr){14-16}
Reference & Replay & IECLA & $\Delta$ & Replay & IECLA & $\Delta$ & Replay & IECLA & $\Delta$ & Replay & IECLA & $\Delta$ & Replay & IECLA & $\Delta$ \\
\midrule
Human Review & 0.9071 & \textbf{0.9315} & \teacherdelta{+0.0244} & 0.9361 & \textbf{0.9537} & \teacherdelta{+0.0176} & 0.7993 & \textbf{0.8478} & \teacherdelta{+0.0485} & 0.2055 & \textbf{0.4320} & \teacherdelta{+0.2265} & 0.1522 & \textbf{0.3601} & \teacherdelta{+0.2079} \\
Qwen Chat & 0.8943 & \textbf{0.9331} & \teacherdelta{+0.0388} & 0.9229 & \textbf{0.9517} & \teacherdelta{+0.0288} & 0.8131 & \textbf{0.8685} & \teacherdelta{+0.0554} & 0.3334 & \textbf{0.5164} & \teacherdelta{+0.1830} & 0.2630 & \textbf{0.4500} & \teacherdelta{+0.1870} \\
Qwen Reasoner & 0.8914 & \textbf{0.9313} & \teacherdelta{+0.0399} & 0.9147 & \textbf{0.9489} & \teacherdelta{+0.0342} & 0.7993 & \textbf{0.8616} & \teacherdelta{+0.0623} & 0.3660 & \textbf{0.5440} & \teacherdelta{+0.1780} & 0.2880 & \textbf{0.4842} & \teacherdelta{+0.1962} \\
GPT Terra & 0.8851 & \textbf{0.9323} & \teacherdelta{+0.0472} & 0.9187 & \textbf{0.9521} & \teacherdelta{+0.0334} & 0.7924 & \textbf{0.8547} & \teacherdelta{+0.0623} & 0.3074 & \textbf{0.5395} & \teacherdelta{+0.2321} & 0.2446 & \textbf{0.4745} & \teacherdelta{+0.2299} \\
VLM mean & 0.8903 & \textbf{0.9322} & \teacherdelta{+0.0419} & 0.9188 & \textbf{0.9509} & \teacherdelta{+0.0321} & 0.8016 & \textbf{0.8616} & \teacherdelta{+0.0600} & 0.3356 & \textbf{0.5333} & \teacherdelta{+0.1977} & 0.2652 & \textbf{0.4695} & \teacherdelta{+0.2043} \\
\bottomrule
\end{tabular}
}%
}
\end{table}

The replay results in Table~\ref{tab:frozen-teacher-replay} show that the learned predictor is not merely a compressed copy of the teacher's ranking. IECLA is higher on every metric for the three individual VLM references and on the VLM mean, while it also improves Human Review agreement on all metrics except the tied Top-5 score. The largest gaps occur in signed measures: on the VLM mean, Signed Spearman rises from 0.4330 to 0.5100, Sign Accuracy from 0.3356 to 0.5333, and Sign F1 from 0.2652 to 0.4695. This pattern is consistent with the predictor learning a more stable signed representation from the intervention supervision rather than reproducing a single noisy replay.

The result also clarifies the role of direct replay. The teacher replay remains the exact intervention-defined source of supervision, but its direct maps are weaker than the learned predictor under the independent VLM and Human Review references. At deployment, replay would require repeated evaluator calls for candidate and matched reference interventions, whereas IECLA reuses one source-emotion inference and applies the learned predictor once per region. Although latency was not measured in this experiment, this replacement reduces counterfactual computation and is expected to lower inference time.

\FloatBarrier
\section{Agent Template}
\label{app:agent-template}

The extended agent retains attribution-grounded local editing while admitting two optional, independently controlled channels for scene atmosphere and minor-object operations. Editable and preservation evidence remain distinct throughout planning. A fourth round audits the complete assembled instruction after the structured plan has passed visual and structural review. New-object proposals are treated as context-conditioned editing hypotheses rather than observed attribution evidence. Figure~\ref{fig:agent-prompt-template} summarizes the paper-level template.

\begin{center}
\centering
\setlength{\fboxsep}{5pt}
\begin{minipage}{0.965\textwidth}
\footnotesize
\centering

\begin{minipage}{0.955\linewidth}
\fcolorbox{black!55}{black!4}{%
\begin{minipage}{\dimexpr\linewidth-2\fboxsep-2\fboxrule\relax}
\textbf{Locked request}\quad
source image $I$; predicted source emotion $y$; target emotion $t$; attributed regions
$r_i=(\texttt{id},\texttt{label},\texttt{bbox},\hat c_i)$ partitioned into editable and preservation evidence; verified placement spaces; output geometry.
Region IDs, boxes, emotion pair, and output size are immutable.
\end{minipage}}
\end{minipage}

\vspace{2pt}$\Downarrow$\vspace{2pt}

\begin{minipage}{0.955\linewidth}
\begin{minipage}[t]{0.455\linewidth}
\fcolorbox{blue!55!black}{blue!5}{%
\parbox[t][7.4em][t]{\dimexpr\linewidth-2\fboxsep-2\fboxrule\relax}{%
\centering\textbf{Round 1A: Evidence verification}\par
\raggedright\vspace{2pt}
Verify editable and preservation evidence, the primary emotional carrier, visible affective cues, scene lighting and palette, existing minor objects, and feasible placement spaces. Report observed evidence and editability risks only.
}}
\end{minipage}
\hfill\raisebox{-3.7em}{$\Longrightarrow$}\hfill
\begin{minipage}[t]{0.455\linewidth}
\fcolorbox{orange!70!black}{orange!7}{%
\parbox[t][7.4em][t]{\dimexpr\linewidth-2\fboxsep-2\fboxrule\relax}{%
\centering\textbf{Round 1B: Object association}\par
\raggedright\vspace{2pt}
Propose up to three context-compatible minor-object hypotheses from verified objects or placement spaces. Prefer \textit{modify} over \textit{remove} over \textit{add}; returning none is valid.\par
\vfill\centering\scriptsize\textit{Editing hypothesis, not attribution evidence}
}}
\end{minipage}
\end{minipage}

\vspace{2pt}$\Downarrow$\vspace{2pt}

\begin{minipage}{0.955\linewidth}
\fcolorbox{teal!60!black}{teal!5}{%
\begin{minipage}{\dimexpr\linewidth-2\fboxsep-2\fboxrule\relax}
\centering\textbf{Round 2: Budgeted route and action planning}\par
\raggedright\vspace{3pt}
\begin{minipage}[t]{0.31\linewidth}
\centering\textbf{Local cue}\par
\raggedright Existing editable content; reduce the source cue and/or establish a visible target cue.
\end{minipage}\hfill
\begin{minipage}[t]{0.31\linewidth}
\centering\textbf{Scene atmosphere}\par
\raggedright Approved photometric or environmental dimensions; preserve scene identity and geometry.
\end{minipage}\hfill
\begin{minipage}[t]{0.31\linewidth}
\centering\textbf{Minor object}\par
\raggedright Modify, add, or remove one verified secondary object under placement and occlusion constraints.
\end{minipage}
\vspace{3pt}\par
\centering\scriptsize Select the least disruptive valid route; use at most one operation per channel.
\end{minipage}}
\end{minipage}

\vspace{2pt}$\Downarrow$\vspace{2pt}

\begin{minipage}{0.955\linewidth}
\fcolorbox{red!55!black}{red!4}{%
\begin{minipage}{\dimexpr\linewidth-2\fboxsep-2\fboxrule\relax}
\textbf{Round 3: Visual and structural critic}\quad
Check executability, emotional completeness, identity and geometry preservation, object existence or placement, occlusion, perspective, and cross-channel consistency. Approve, minimally revise, drop an optional channel, or mark the task invalid.
\end{minipage}}
\end{minipage}

\vspace{3pt}
\begingroup
\setlength{\fboxsep}{3pt}
\begin{minipage}{0.955\linewidth}
\begin{minipage}[t]{0.205\linewidth}
\fcolorbox{black!50}{black!3}{\parbox[c][6.2em][c]{\dimexpr\linewidth-2\fboxsep-2\fboxrule\relax}{\centering\textbf{Draft compiler}\par\vspace{2pt}Assemble approved actions into an ordered draft.}}
\end{minipage}
\hfill$\Longrightarrow$\hfill
\begin{minipage}[t]{0.205\linewidth}
\fcolorbox{violet!65!black}{violet!5}{\parbox[c][6.2em][c]{\dimexpr\linewidth-2\fboxsep-2\fboxrule\relax}{\centering\textbf{Round 4 auditor}\par\vspace{2pt}Check emotion logic, referents, conflicts, action load, and route consistency.}}
\end{minipage}
\hfill$\Longrightarrow$\hfill
\begin{minipage}[t]{0.205\linewidth}
\fcolorbox{black!50}{black!3}{\parbox[c][6.2em][c]{\dimexpr\linewidth-2\fboxsep-2\fboxrule\relax}{\centering\textbf{Final compiler}\par\vspace{2pt}Validate locked fields and apply one protection profile.}}
\end{minipage}
\hfill$\Longrightarrow$\hfill
\begin{minipage}[t]{0.205\linewidth}
\fcolorbox{blue!60!black}{blue!7}{\parbox[c][6.2em][c]{\dimexpr\linewidth-2\fboxsep-2\fboxrule\relax}{\centering\textbf{Image editor}\par\vspace{2pt}Apply the audited instruction to the source image.}}
\end{minipage}
\end{minipage}
\endgroup

\vspace{4pt}
\begin{minipage}{0.955\linewidth}
\fcolorbox{black!40}{yellow!8}{%
\begin{minipage}{\dimexpr\linewidth-2\fboxsep-2\fboxrule\relax}
\textbf{Control invariants.}
One primary emotional action; at most one atmosphere action and one minor-object action. Protect regions identified as preservation evidence, people, and primary objects; preserve the selected route's structural invariants; and keep attribution evidence distinct from proposed additions. Invalid plans are not silently replaced by a native prompt or unchanged input.
\end{minipage}}
\end{minipage}
\end{minipage}


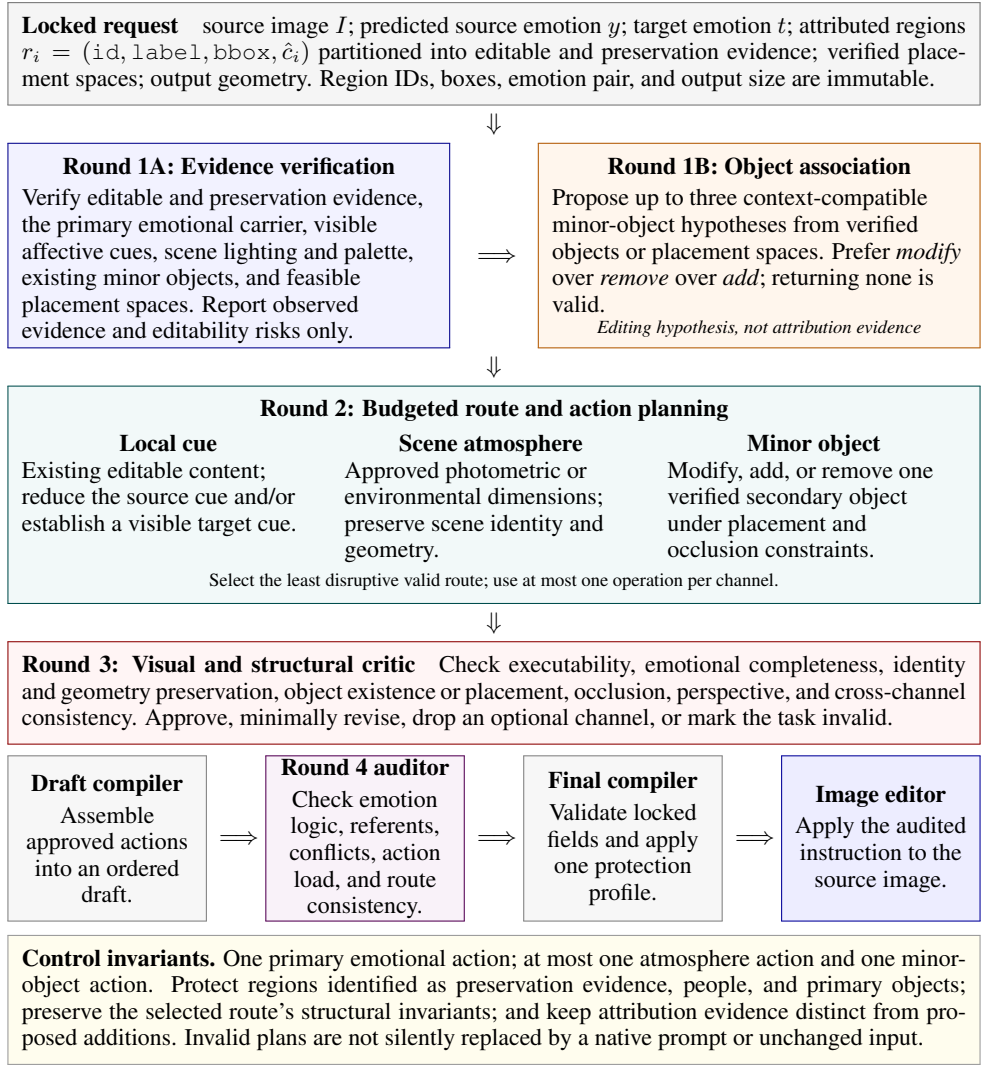
\captionof{figure}{Extended attribution-grounded editing agent template. Verified evidence is routed through local, atmosphere, and optional minor-object operations; structural review and whole-instruction auditing constrain the final editor instruction.}
\label{fig:agent-prompt-template}
\end{center}

The route-specific compiler emits only approved clauses and one compatible protection profile. This avoids contradictory constraints such as simultaneously adding an object and requiring object count to remain unchanged.

\section{Implementation Details}
\label{app:implementation-details}

This appendix specifies the training configuration and editing controls used in Sections~\ref{sec:method} and~\ref{sec:experimental-setup}.

\subsection{Data Provenance and Splits}

The contribution-training, locked attribution-test, and editing-evaluation pools are constructed from EmoVerse. Images belonging to the external evaluation datasets, including EmoSet and EmoArt, are removed before contribution-model training. The training pool supplies 1,056,949 candidate masks, whereas the image-disjoint locked attribution test contains 7,278 images and 53,180 masks. The editing benchmark uses a separate set of 680 challenging images selected to include multi-subject scenes, scenes without a dominant subject, and scenes containing many objects. Deduplication and overlap checks are applied before the final splits, leaving the training, locked attribution-test, and editing-evaluation pools mutually disjoint.

\subsection{Editing Tasks and Evaluation Metrics}

\paragraph{Task construction.}
The main editing pool contains 680 source images, each paired with all eight target emotions: amusement, anger, awe, contentment, disgust, excitement, fear, and sadness. This construction yields $680\times8=5{,}440$ source--target task keys, including 680 same-emotion tasks $(t=y)$ and 4,760 cross-emotion tasks $(t\neq y)$, where the frozen image-level evaluator supplies the source emotion $y$. The six editors use the same task keys, and Native and IECLA-guided conditions share the source image, target emotion, seed, sampler, and resolution rule. Tables~\ref{tab:editing-native}, \ref{tab:system-ablation}, and~\ref{tab:loss-ablation} use the same evaluation set for their corresponding Qwen-Edit conditions, which accounts for their identical shared entries.

Cross-dataset editing applies every target emotion to every image used in the attribution evaluation. The 3,370 external images therefore produce 26,960 editing tasks, comprising 3,370 same-emotion and 23,590 cross-emotion tasks. Table~\ref{tab:cross-dataset} averages over 4,896 COCO tasks, 7,320 EmoSet tasks, 12,352 FI tasks, 1,808 Emotion6 tasks, and 584 EmoArt tasks. The stability analysis in Appendix~\ref{app:stability} is a separate diagnostic: its 16 images are edited only toward the other seven emotions, giving 112 tasks for each seed and resolution.

\begin{table}[!ht]
\centering
\appendixtablesetup
\caption{Editing-task accounting. The main and cross-dataset evaluations include all eight targets, whereas the controlled cross-emotion diagnostics exclude $t=y$.}
\label{tab:editing-task-accounting}
\begin{tabular}{lrrrrr}
\toprule
Evaluation & Images & Targets/image & Same-emotion & Cross-emotion & Tasks \\
\midrule
Main editing pool & 680 & 8 & 680 & 4,760 & 5,440 \\
COCO & 612 & 8 & 612 & 4,284 & 4,896 \\
EmoSet & 915 & 8 & 915 & 6,405 & 7,320 \\
FI & 1,544 & 8 & 1,544 & 10,808 & 12,352 \\
Emotion6 & 226 & 8 & 226 & 1,582 & 1,808 \\
EmoArt & 73 & 8 & 73 & 511 & 584 \\
\midrule
\textbf{Cross-dataset total} & \textbf{3,370} & \textbf{8} & \textbf{3,370} & \textbf{23,590} & \textbf{26,960} \\
\midrule
Stability, per seed/resolution & 16 & 7 & 0 & 112 & 112 \\
\bottomrule
\end{tabular}
\end{table}

\paragraph{Affective metrics.}
Let $I_s$ and $I_e$ denote the source and edited images, and let $t$ be the requested emotion. The frozen EmoVerse-supervised evaluator $E$ defines
\begin{equation}
\begin{aligned}
&\operatorname{Success}(I_e,t)
 =\mathbb{I}\!\left[\arg\max_{e}E_e(I_e)=t\right],\\
&\Delta p(I_s,I_e,t)
 =E_t(I_e)-E_t(I_s).
\end{aligned}
\label{eq:editing-target-metrics}
\end{equation}
An independent CLIP-based emotion classifier $C$ provides a second evaluation:
\begin{equation}
\begin{aligned}
&\operatorname{Emo\text{-}A}(I_e,t)
 =\mathbb{I}\!\left[\arg\max_{e}C_e(I_e)=t\right],\\
&\operatorname{Emo\text{-}S}(I_s,I_e,t)
 =C_t(I_e)-C_t(I_s).
\end{aligned}
\label{eq:editing-emo-metrics}
\end{equation}
The classifier $C$ follows the EmoEdit evaluation model: a frozen CLIP ViT-B/32 image encoder feeds a two-layer classifier with dimensions $512\rightarrow256\rightarrow8$ and a softmax output. Its released checkpoint was pretrained on EmoSet. We performed image-level overlap checks between its EmoSet training data and the EmoSet subset used in our cross-dataset evaluation, confirming that the two sets are disjoint.
Thus, Target Success and Emo-A are target-class accuracies from different held-out evaluators, while $\Delta p$ and Emo-S measure target-probability changes from the corresponding evaluators. Reported entries are arithmetic means over the task set specified for each experiment; source-image cluster resampling is used when confidence intervals are reported.

\paragraph{Evaluator roles and metric provenance.}
The frozen evaluator $E$ provides the source-emotion label and counterfactual supervision, while the contribution predictor scores candidate regions and the target-aware agent plans edits from those scores. The classifier $C$ is used only for post-hoc Emo-A and Emo-S evaluation; its scores are never fed back to the predictor or agent. VLM references, Human Review, and the User Study provide complementary external and perceptual evaluations.

\paragraph{Global preservation metrics.}
All source and edited images are converted to RGB and resized to $512\times512$ with Lanczos interpolation. PSNR, SSIM, and LPIPS compare the complete images, while CLIP-I is the cosine similarity between their CLIP image embeddings. In Tables~\ref{tab:system-ablation} and~\ref{tab:loss-ablation}, the prefix G. denotes the same complete-image evaluation. G. SSIM is computed on uint8 images with data range 255. G. LPIPS uses the AlexNet LPIPS backbone on float32 images scaled to $[0,1]$. We call the official LPIPS implementation with \texttt{normalize=True}, which internally maps these inputs to the expected $[-1,1]$ range:
\begin{equation}
\operatorname{G.LPIPS}=D_{\mathrm{LPIPS}}(I_s,I_e).
\label{eq:global-lpips}
\end{equation}
All compared conditions use this identical normalized implementation. The reported values therefore follow the standard LPIPS preprocessing while retaining the common $[0,1]$ image representation used by the rest of our evaluation code.

\paragraph{Common outside-region protocol.}
Outside-region metrics measure changes beyond a common allowed editing area. For each task, the selected binary SAM masks from all compared conditions are resized to $512\times512$ by nearest-neighbor interpolation and combined by a hard union,
\begin{equation}
A=\bigcup_{v\in\mathcal{V}}M_v,\qquad O=\neg A,
\label{eq:common-outside-region}
\end{equation}
where $\mathcal{V}$ is the set of variants in the comparison. Fixing $A$ across variants prevents a condition from improving its preservation score by declaring a larger editable region. In the loss-function ablation, $\mathcal{V}$ contains the complete objective and the four loss-removed variants. Their common allowed region covers 22.7521\% of the image on average, leaving 77.2479\% for outside-region evaluation. This experiment uses the same 680 source images and all eight target emotions, yielding $680\times8=5{,}440$ tasks.

O. MAE averages the RGB absolute error over $O$:
\begin{equation}
\operatorname{O.MAE}
=\frac{1}{|O|}\sum_{p\in O}
\frac{\|I_e(p)-I_s(p)\|_1}{3}.
\label{eq:outside-mae}
\end{equation}
O. SSIM first computes the full-resolution SSIM map and then averages its values over $O$. For O. LPIPS, the allowed region is restored to the source image,
\begin{equation}
I_{\mathrm{outside}}
=A\odot I_s+(1-A)\odot I_e,\qquad
\operatorname{O.LPIPS}
=D_{\mathrm{LPIPS}}(I_s,I_{\mathrm{outside}}).
\label{eq:outside-lpips}
\end{equation}
O. LPIPS is therefore an outside-focused perceptual distance rather than a pixelwise regional average; convolutional receptive fields may still cross the mask boundary.

The affective and preservation metrics are interpreted jointly: minimal image change is not beneficial when the requested emotion is not expressed, while improved affective scores accompanied by poor outside-region preservation indicate unnecessary rewriting.

\subsection{Counterfactual Reference Calibration}

The reference correction in Equation~\ref{eq:counterfactual-contribution} uses dataset-level calibration rather than a fixed number $k$ of reference masks for each candidate. We sample $N_{\mathrm{ref}}=10{,}000$ training records and generate one reference intervention per record. Reference regions are matched to candidate regions in nominal area, spatial position, and shape. The resulting reference effects are averaged within source-emotion and area bins, using five area ranges: $[0,0.005)$, $[0.005,0.02)$, $[0.02,0.10)$, $[0.10,0.30)$, and $[0.30,1]$. Each emotion--area cell contains at least 30 references; otherwise, the area-level and then global mean provide the fallback. Thus, $\mathbb{E}[\Delta_r]$ denotes the corresponding calibration mean, not a per-region Monte Carlo average.

Both candidate and reference interventions replace the matched region with a radius-24 Gaussian-blurred version. Matching area, position, and shape ensures comparable intervention geometry, while averaging over the calibration set estimates the blur-induced nuisance effect. Subtracting this expectation leaves the candidate-specific residual as the corrected contribution.

We interpret $c_i$ on this calibrated relative scale: $c_i<0$ indicates an effect below the matched-reference mean, while its role as an opposing editing cue is validated against Human Review.

\subsection{Predictor Training and Objective}

\paragraph{Predictor training.}
The contribution predictor is trained on 1,056,949 candidate masks and validated on 58,591 disjoint masks. Every loss term uses the corrected counterfactual contribution $c_i$ from Equation~\ref{eq:counterfactual-contribution} as its supervision target, and checkpoint selection minimizes validation MAE on these corrected contributions. Training uses three NVIDIA A800 GPUs.

\paragraph{Supervision and weighted regression.}
Let $\widehat c_i$ denote the predicted contribution and $\tau=0.1$ the fixed sign threshold. The target sign is negative for $c_i<-\tau$, neutral for $|c_i|\leq\tau$, and positive for $c_i>\tau$. We define the unit-transition Huber penalty as
\begin{equation}
\rho(e)=
\begin{cases}
\tfrac{1}{2}e^2, & |e|\leq 1,\\
|e|-\tfrac{1}{2}, & |e|>1.
\end{cases}
\label{eq:appendix-huber-penalty}
\end{equation}
Let $b_i$ be the evaluator's pre-intervention class and $\widetilde y_i$ the image-level training label. The sample weight
\begin{equation}
w_i=3^{\mathbb{I}[c_i<-\tau]}
1.5^{\mathbb{I}[|c_i|>\tau]}
0.5^{\mathbb{I}[b_i\neq\widetilde y_i]}
\label{eq:appendix-sample-weight}
\end{equation}
combines the three multipliers rather than selecting only one of them. The normalized regression term is
\begin{equation}
\mathcal{L}_{\mathrm{Huber}}^{w}
=\frac{\sum_i w_i\rho(\widehat c_i-c_i)}{\sum_i w_i}.
\label{eq:appendix-weighted-huber}
\end{equation}

\paragraph{Within-image ranking.}
Let $g(i)$ identify the source image of region $i$. Absolute and signed ranking use, respectively, the valid pair sets
\begin{equation}
\begin{aligned}
&\mathcal{P}_{\mathrm{abs}}=\{(i,j):g(i)=g(j),\ i<j,\ |c_i|\neq|c_j|\},\\
&\mathcal{P}_{\mathrm{signed}}=\{(i,j):g(i)=g(j),\ i<j,\ c_i\neq c_j\}.
\end{aligned}
\label{eq:appendix-ranking-pairs}
\end{equation}
With $\sigma(\cdot)$ denoting the sigmoid function, the two losses are
\begin{equation}
\begin{aligned}
&\mathcal{L}_{\mathrm{abs}}
 =-\frac{1}{|\mathcal{P}_{\mathrm{abs}}|}
 \sum_{(i,j)\in\mathcal{P}_{\mathrm{abs}}}
 \log\sigma\!\left(s_{ij}^{a}(|\widehat c_i|-|\widehat c_j|)\right),~
 s_{ij}^{a}=\operatorname{sign}(|c_i|-|c_j|),\\
&\mathcal{L}_{\mathrm{signed}}
 =-\frac{1}{|\mathcal{P}_{\mathrm{signed}}|}
 \sum_{(i,j)\in\mathcal{P}_{\mathrm{signed}}}
 \log\sigma\!\left(s_{ij}^{s}(\widehat c_i-\widehat c_j)\right), ~
 s_{ij}^{s}=\operatorname{sign}(c_i-c_j).
\end{aligned}
\label{eq:appendix-ranking-losses}
\end{equation}
The absolute term orders regions by contribution strength, whereas the signed term preserves their ordering on the corrected contribution scale.

\paragraph{Sign and affective auxiliary supervision.}
For the three sign classes $q_i\in\{-,0,+\}$, let $p_i^{q_i}$ be the predicted probability of the target class. With class weights $(\alpha_{-},\alpha_{0},\alpha_{+})=(4.0,0.3,1.1)$ and focal exponent $\gamma=2$, the sign-classification loss is
\begin{equation}
\mathcal{L}_{\mathrm{focal}}
=-\frac{\sum_i\alpha_{q_i}(1-p_i^{q_i})^2\log p_i^{q_i}}
{\sum_i\alpha_{q_i}}.
\label{eq:appendix-focal-loss}
\end{equation}
The auxiliary heads predict the intensity difference $d_i^{\mathrm{int}}$ and the three-dimensional VAD difference $\mathbf d_i^{\mathrm{VAD}}$ associated with the same intervention. For a batch of $B$ regions, their loss is
\begin{equation}
\begin{aligned}
\mathcal{L}_{\mathrm{aux}}={}&
\frac{1}{B}\sum_i\rho(\widehat d_i^{\mathrm{int}}-d_i^{\mathrm{int}})
+\frac{0.2}{3B}\sum_i\sum_{k=1}^{3}
\rho(\widehat d_{ik}^{\mathrm{VAD}}-d_{ik}^{\mathrm{VAD}}).
\end{aligned}
\label{eq:appendix-auxiliary-loss}
\end{equation}
The coefficient $0.2$ applies within $\mathcal{L}_{\mathrm{aux}}$, Equation~\ref{eq:training-objective} further weights the full auxiliary term by $0.10$.

\paragraph{Grouped sampling and pair construction.}
Candidate regions are grouped by source image, image groups are shuffled each epoch, and complete groups are packed without splitting into batches of up to 128 regions per process. With three training processes, this gives the effective batch size of 384 reported above. Both ranking terms enumerate every valid unordered pair within each image group; they use neither random pair subsampling nor a ranking margin. Tied targets are omitted, and image groups with fewer than two regions contribute zero to the ranking terms.

\subsection{Attribution Baseline Implementations}

The vision-encoder contribution baselines are trained comparison versions used to produce signed heatmaps. They share the candidate masks, counterfactual supervision, and mask-to-pixel rendering protocol of IECLA, but use simpler backbone-specific predictors. The CLIP version combines frozen ViT-L/14 region-crop and full-image features with emotion conditioning, object labels, and region metadata. The SigLIP2 version uses a frozen SigLIP2-base encoder with region and full-image context, emotion and object conditions, and the same metadata. The DINOv3 version pools frozen ViT-7B/16 patch tokens within each SAM mask and combines the region representation with full-image context, emotion and object conditions, and metadata. Each predictor outputs a signed regional contribution that is projected onto the fixed masks; these baselines are trained specifically for heatmap comparison and are not native attention maps.

Grad-CAM, Grad-CAM++, HiResCAM, LayerCAM, EigenCAM, and Ablation-CAM use the frozen eight-class emotion evaluator $E$ as their target network and explain its inferred source-emotion class. RISE-256 uses the same evaluator with 256 random masks sampled on a $7\times7$ grid. Each resulting saliency map is resized to the input resolution, and its score for a candidate region is the mean saliency within the corresponding fixed SAM mask. Direct attention instead uses the CLIP visual encoder from the CLIP contribution baseline: multi-head self-attention is augmented with the identity connection and rolled out across transformer layers to obtain the CLS-to-patch map. ADAPT-TF likewise follows its own Qwen image-token attention path rather than the evaluator $E$. These heatmap methods produce nonnegative region scores and are therefore evaluated only with unsigned ranking and localization metrics.

\paragraph{Agent constraints and candidate selection.}
The predictor supplies up to three attributed regions with $\widehat c_i>0.05$ as editable evidence and retains strongly negative regions as preservation evidence. The agent verifies a primary emotional carrier, and the local action remains anchored to the editable evidence. Scene-atmosphere changes and minor-object operations are optional, independently budgeted channels that require visible scene support; a proposed added object is recorded as an editing hypothesis rather than attributed evidence. At most one primary, one atmosphere, and one minor-object operation are retained. A visual and structural critic checks feasibility and preservation, after which a deterministic draft compiler and integrated-instruction auditor instantiate the template in Appendix~\ref{app:agent-template}. Reporting evaluators remain held out from planning and candidate selection.

\paragraph{Agent VLM configuration.}
All agent rounds use the same locally served vision-language model under the API identifier \texttt{qwen36-vision}. The model is loaded from \nolinkurl{Qwen3.6-35B-A3B-Uncensored-HauhauCS-Aggressive-Q8_K_P.gguf} together with \nolinkurl{mmproj-Qwen3.6-35B-A3B-Uncensored-HauhauCS-Aggressive-f16.gguf}. It uses the \texttt{qwen35moe} architecture with 34.66B total parameters and approximately 3B active parameters, and the language checkpoint uses \texttt{Q8\_K\_P} quantization. Thinking is enabled for every round; decoding uses temperature 0.1, the server default top-$p$ of 0.95, at most 1,024 output tokens per round, and an 8,192-token service context.

The original image is transmitted as base64 data without client-side resizing. The vision encoder has a nominal input size of 768 and a patch size of 16, accepts dynamic inputs from 8,192 to 4,194,304 pixels, and uses a \texttt{qwen3vl\_merger} projector with spatial merge size 2.

\paragraph{Editor configurations.}
The editing experiments use Qwen-Edit, InstructPix2Pix, FLUX.2, LongCat, OmniGen2, and HiDream. Qwen-Edit denotes Qwen-Image-Edit-2511, FLUX.2 denotes FLUX.2 Klein 4B, LongCat denotes LongCat-Turbo, and HiDream denotes HiDream-E1.1; InstructPix2Pix and OmniGen2 use their official released checkpoints. The stability analysis in Appendix~\ref{app:stability} runs Qwen-Edit with Lightning four-step sampling.

\paragraph{Editing targets.}
The eight target emotions are amusement, anger, awe, contentment, disgust, excitement, fear, and sadness. Native and IECLA-guided outputs share the source image, target emotion, seed, sampler, and resolution rule within each editor. Absolute scores across different backbones are not interpreted as a ranking of editors.

\end{document}